\documentclass[acmtog]{acmart}
\acmSubmissionID{382}
\usepackage{tabularx}
\usepackage{siggraph}
\acrodef{hsi}[HSI]{Human-Scene Interaction}
\acrodef{ik}[IK]{Inverse Kinematic}
\acrodef{hoi}[HOI]{Human-Object Interaction}
\acrodef{dataset}[\texttt{LINGO}]{\underline{T}racking H\underline{um}an \underline{A}ctio\underline{n}s in \underline{S}cenes}
\acrodef{cvae}[cVAE]{conditional Variational Auto-Encoder}
\acrodef{icp}[ICP]{Iterative Closest Points}
\acrodef{mocap}[MoCap]{motion-captured}
\acrodef{vit}[ViT]{Vision Transformer}
\acrodef{fid}[FID]{Fréchet Inception Distance}

\usepackage{array}

\copyrightyear{2024}
\acmYear{2024}
\setcopyright{acmlicensed}\acmConference[SA Conference Papers '24]{SIGGRAPH Asia
2024 Conference Papers}{December 3--6, 2024}{Tokyo, Japan}
\acmBooktitle{SIGGRAPH Asia 2024 Conference Papers (SA Conference Papers '24),
December 3--6, 2024, Tokyo, Japan}
\acmDOI{10.1145/3680528.3687595}
\acmISBN{979-8-4007-1131-2/24/12}

\acrodef{hsi}[HSI]{Human-Scene Interaction}
\acrodef{ik}[IK]{Inverse Kinematic}
\acrodef{hoi}[HOI]{Human-Object Interaction}
\acrodef{dataset}[\texttt{LINGO}]{\underline{T}racking H\underline{um}an \underline{A}ctio\underline{n}s in \underline{S}cenes}
\acrodef{cvae}[cVAE]{conditional Variational Auto-Encoder}
\acrodef{icp}[ICP]{Iterative Closest Points}
\acrodef{mocap}[MoCap]{motion-captured}
\acrodef{vit}[ViT]{Vision Transformer}
\acrodef{fid}[FID]{Fréchet Inception Distance}
\begin{document}
\title{Autonomous Character-Scene Interaction Synthesis from Text Instruction}

\author{Nan Jiang}
\orcid{0009-0006-5726-7672}
\affiliation{%
 \institution{Institute for AI, Peking University}
 \country{China}
}
\affiliation{%
 \institution{National Key Lab of General AI, BIGAI}
 \country{China}
}
\email{nan.jiang@stu.pku.edu.cn}
\authornote{Both authors contributed equally to this research.}

\author{Zimo He}
\orcid{0009-0005-2330-559X}
\affiliation{%
 \institution{Institute for AI, Peking University}
 \country{China}
}
\email{milleret@stu.pku.edu.cn}
\authornotemark[1]

\author{Zi Wang}
\orcid{0009-0003-3327-7700}
\affiliation{%
 \institution{Beijing University of Posts and Telecommunications}
 \country{China}
}
\affiliation{%
 \institution{National Key Lab of General AI, BIGAI}
 \country{China}
}
\email{wangzi1@bupt.edu.cn}

\author{Hongjie Li}
\orcid{0009-0000-0695-1008}
\affiliation{%
 \institution{Institute for AI, Peking University}
 \country{China}
}
\email{lihongjie@stu.pku.edu.cn}

\author{Yixin Chen}
\orcid{0000-0002-8176-0241}
\affiliation{%
 \institution{National Key Lab of General AI, BIGAI}
 \country{China}
}
\email{ethanchen@g.ucla.edu}

\author{Siyuan Huang}
\orcid{0000-0003-1524-7148}
\affiliation{%
 \institution{National Key Lab of General AI, BIGAI}
 \country{China}
}
\email{huangsiyuan@ucla.edu}
\authornote{Corresponding authors}

\author{Yixin Zhu}
\orcid{0000-0001-7024-1545}
\affiliation{%
 \institution{Institute for AI, Peking University}
 \country{China}
}
\email{yixin.zhu@pku.edu.cn}
\authornotemark[2]

\begin{abstract}
Synthesizing human motions in 3D environments, particularly those with complex activities such as locomotion, hand-reaching, and \ac{hoi}, presents substantial demands for user-defined waypoints and stage transitions. These requirements pose challenges for current models, leading to a notable gap in automating the animation of characters from simple human inputs. This paper addresses this challenge by introducing a comprehensive framework for synthesizing multi-stage scene-aware interaction motions directly from a single text instruction and goal location. Our approach employs an auto-regressive diffusion model to synthesize the next motion segment, along with an autonomous scheduler predicting the transition for each action stage. To ensure that the synthesized motions are seamlessly integrated within the environment, we propose a scene representation that considers the local perception both at the start and the goal location. We further enhance the coherence of the generated motion by integrating frame embeddings with language input. Additionally, to support model training, we present a comprehensive \ac{mocap} dataset comprising 16 hours of motion sequences in 120 indoor scenes covering 40 types of motions, each annotated with precise language descriptions. Experimental results demonstrate the efficacy of our method in generating high-quality, multi-stage motions closely aligned with environmental and textual conditions. Project page: \href{https://lingomotions.com}{\color{magenta}{https://lingomotions.com}}
\end{abstract}



\begin{teaserfigure}
    \includegraphics[width=\linewidth]{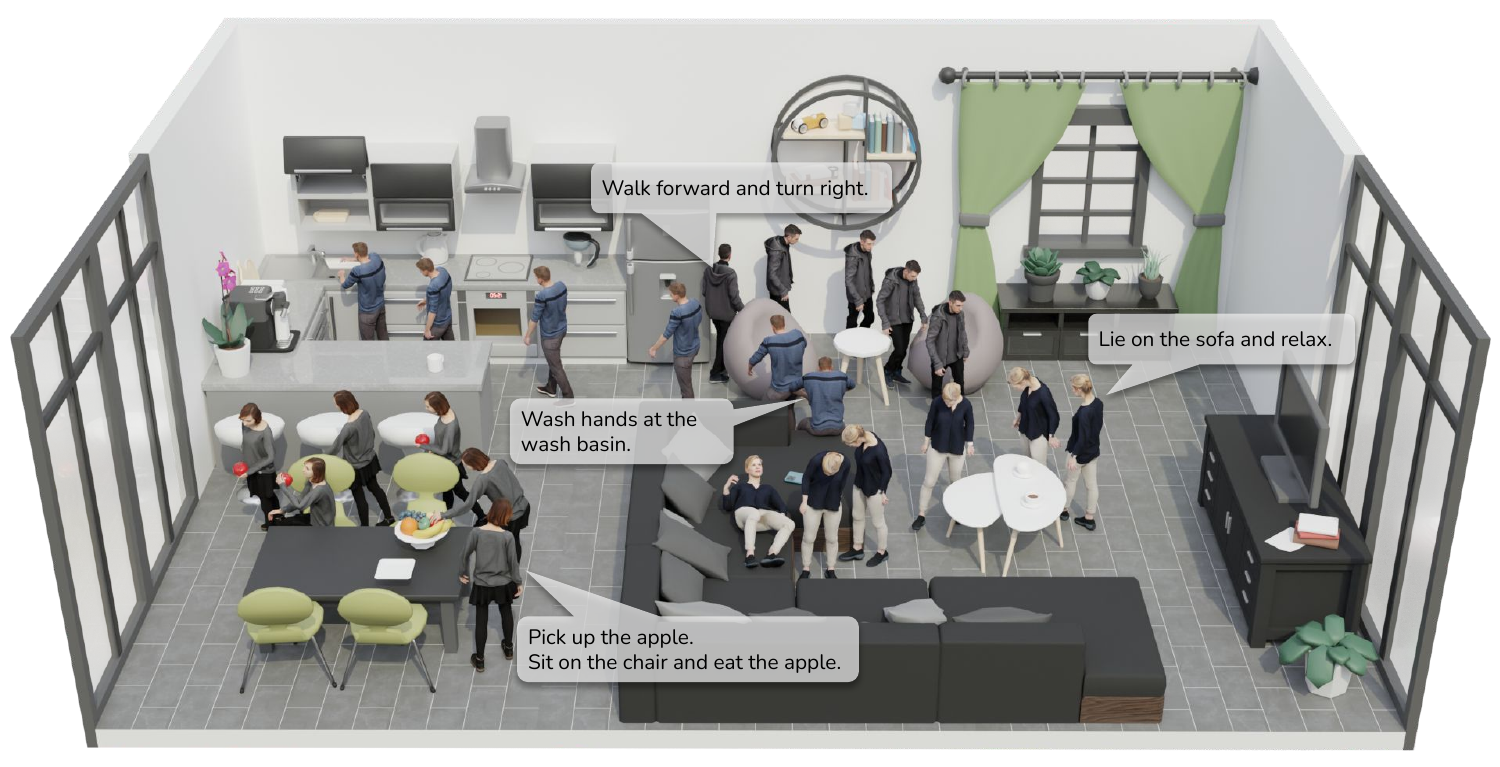}
    \caption{\textbf{Autonomous \acs{hsi} synthesis}. Our proposed method generates realistic character motion in 3D scenes based on a single textual instruction and goal location, incorporating seamless transitions between locomotion and \acs{hoi} autonomously.}
    \label{fig:teaser}
\end{teaserfigure}

\maketitle

\section{Introduction}

Language-guided character motion synthesis within dynamic 3D environments presents a profound challenge in addressing the complexity of multi-stage interactions such as locomotion, hand reaching, and \ac{hoi}. Unlike humans, who effortlessly interpret and respond to verbal instructions in varied contexts, current motion synthesis methods often fall short of replicating this intuitive capability. Despite recent advances in motion synthesis methods, several challenges remain unaddressed. 

Firstly, a primary obstacle in this field is the absence of a unified framework that integrates the various stages of \acp{hsi} into a single pipeline. Instead, different stages of motion synthesis, such as walking, reaching, or interacting with objects, are often modeled by separate specialized systems \cite{liu2018learning,liu2017learning,merel2020catch, hassan2023synthesizing}. This fragmented approach results in a lack of coherence in the synthesized motions, making it difficult to sustain long-term interaction sequences that are contextually aligned to the input. Secondly, another significant gap lies in the complexity of the input data required. Some recent works rely on additional inputs that specify the trajectories of objects \cite{li2023controllable}, keypoints of humans \cite{jiang2024scaling}, or motion phase labelling \cite{starke2019neural}. These dependencies constraint the flexibility of the methods and limit their practical deployment. Finally, the absence of scene-level, well-annotated datasets with text labels poses another critical challenge. Although several datasets \cite{taheri2020grab,bhatnagar22behave,jiang2023full,hassan2019resolving} have recently been proposed, they have failed to cover all spectrums of character-scene interaction, such as grasping, context-rich HOI, and complex scene constraints. Such datasets are essential for developing models that accurately interpret and execute comprehensive motion instructions.

In this paper, we tackle the intricate task of language-guided synthesis of multistage \acp{hsi}, directly confronting the challenges outlined earlier. We aim to reduce the dependency on extra, user-provided input data, aiming to autonomously synthesize motion from a single text instruction and goal location as control signals. Furthermore, we integrate the disjointed processes of locomotion, hand reaching, and \ac{hoi} into a seamless pipeline. This unified approach ensures not only the realism of the motion synthesis but also its functional coherence.

To address the challenge of synthesizing multi-stage \acp{hsi}, our method employs an auto-regressive diffusion model designed to generate subsequent motion segments and an autonomous scheduler predicting the transition timing between stages. Recognizing the need for minimal user inputs, which consist only of a text description and a goal location, we leverage the stage-specific goal encoder to produce appropriate condition terms relative to the goal of the current segment. More specifically, we introduce a dual voxel scene encoder that captures detailed context from both the starting and target positions of the motion segment. For precise and time-specific semantic guidance, we propose integrating the time frame embedding with language embeddings to process textual inputs. This dual representation strategy ensures that the synthesized motions are contextually accurate and highly integrated with the surrounding environment, enhancing realism and applicability.

To bridge the gap between scene, motion, and language, we present LINGO (Language-annotated INteraction, Grasping, and lOcomotion), a comprehensive \ac{mocap} dataset that unifies \ac{hsi} activities. LINGO exceeds prior efforts by alleviating the significant labor in real-world motion capturing through a VR-assisted setup, where the vision of synthetic scenes is projected into a VR headset worn by the motion actor. LINGO provides fully text-annotated, long-term human motions and dynamic object interactions of 16 hours within 120 diverse scenes spanning 40 types of diverse motions.

The primary contributions of this work are threefold. First, we propose a framework synthesizing multi-stage scene-aware human motions autonomously and directly from the text instruction and goal location. Second, our model combines the auto-regressive diffusion model with a novel 3D scene representation and a joint time frame and language embedding, achieving seamless integration between human motions and 3D physical environments. Third, we introduce a comprehensive language-annotated \ac{mocap} dataset featuring \ac{hsi} motions.

\section{Related Work}

The field of \ac{hsi} synthesis has seen significant progress in recent years, with two main research directions: interacting with static scenes and dynamic objects.

\subsection{Interaction Synthesis in Static Scenes}

Regarding the static scenes, early work in this area primarily focused on synthesizing single-frame human poses given a 3D scene configuration \cite{li2019putting,zhang2020generating,zhangsiwei2020generating}. \citet{zhao2022compositional} leveraged contact priors to generate human poses following text conditions. For motion synthesis with static objects, most works have focused on producing motions of locomotion, sitting, and lying with large furniture from pre-defined milestones \cite{wang2021synthesizing,wang2021scene,jiang2024scaling,zhao2023synthesizing}, or tackling short-term motion generation \cite{wang2022humanise,huang2023sceneDiffuser}.
Reinforcement learning-based methods have also been developed. \citet{WanderingsofOdysseusIn3DScenes} employs generative motion primitives and a policy network that enables goal-oriented locomotion. \citet{zhao2023synthesizing} aims to allow digital humans to perform interactive actions such as sitting on a chair or lying on a sofa. UniHSI \cite{xiao2024unified} introduces a unified framework for multiple types of \ac{hsi}. However, the framework's reliance on a contact-based driver guided by a large language model limits its applicability. HUMANISE \cite{wang2022humanise} was the first to explore text-conditioned scene interactions but relied entirely on short synthetic sequences for training, which was expanded by \citet{yi2024generating}. GOAL \cite{taheri2022goal} and SAGA \cite{wu2022saga} generate full-body poses that aim to reach a specific object. However, these models often specialize in limited interaction types, failing to capture detailed semantics from the instruction and geometric constraints of 3D scenes. This limits their applicability for animating realistic characters for desired interactions.

\subsection{Interaction Synthesis for Dynamic Objects}

Another line of work focuses on synthesizing human motion with dynamically involved scene objects. Reinforcement learning has been widely used to learn different skills. Early work simplified the object manipulation problem by explicitly attaching an object to the character's hands \cite{coros2010generalized,peng2019mcp,mordatch2012discovery}, avoiding the grasping process. Other methods have been proposed to train task-specific policies, such as basketball dribbling \cite{liu2018learning}, skateboarding \cite{liu2017learning}, and box manipulation \cite{merel2020catch,hassan2023synthesizing}. \citet{lee2023locomotion} built a unified framework to encompass a series of everyday motions, including locomotion, scene interaction, and object manipulation, similar to our goal. However, it is driven by designated interaction cues, limiting motion diversity and flexibility. Some work tackles interaction motion synthesis without physical simulators\cite{starke2019neural,cui2024anyskill}. \citet{starke2019neural} automates character movements and interactions with objects and uses neural networks to update the state of action dynamically. However, the method faces difficulty in generalizing to acyclic motions. IMoS \cite{ghosh2022imos} synthesizes human and object motions simultaneously after grasping an object, and \citet{li2023object} proposed a framework that synthesizes human motion given object pose trajectories. The most similar works to ours are TRUMANS \cite{jiang2024scaling}, which is based on frame-wise action labels rather than text instructions, and \citet{li2023controllable}, which neglects fine-grained scene geometrical constraints. Furthermore, both require predefined waypoints for humans or objects, making them incapable of being applied to an autonomous digital character.

\subsection{Character-scene Interaction Synthesis Datasets}

\ac{hsi} datasets can be broadly classified into two categories: those that focus on human interactions with objects \cite{taheri2020grab,bhatnagar22behave,jiang2023full} and those that capture human motion within a scene \cite{monszpart2019imapper,savva2016pigraphs,hassan2019resolving,jiang2024scaling}.

A small number of datasets have been developed to capture human manipulation of small objects \cite{taheri2020grab,fan2023arctic} or interactions with large objects \cite{bhatnagar22behave,jiang2023full,zhang2022couch}. The BEHAVE dataset \cite{bhatnagar22behave}, for instance, contains long sequences of humans interacting with 20 types of large objects. The GRAB dataset \cite{taheri2020grab}, on the other hand, focuses on full-body motions of small object manipulation using hands. The CHAIRS \cite{jiang2023full} and COUCH \cite{zhang2022couch} datasets specifically target human interactions with sittable objects. However, a common limitation of these datasets is that they only consider the object during the interaction, disregarding the surrounding environment.

Early attempts to record human activities in a scene rely on multi-view camera setups and reconstruct human motion through key-point detection \cite{monszpart2019imapper,savva2016pigraphs,hassan2019resolving}. These datasets typically feature a limited set of actions, such as sitting, lying, and locomotion. Later, researchers begin to employ \ac{mocap} equipment, including IMU-based and optic-based systems like VICON. The SAMP dataset \cite{hassan2021stochastic}, for example, covers various sitting, lying down, walking, and running styles in a scene. \citet{guzov23ireplica} tracks human movements in large indoor scenes, including interactions with articulated objects. Another notable dataset is CIRCLE \cite{araujo2023circle}, which captures reaching motions to specific locations in cluttered scenes. In CIRCLES, the scene is virtually presented to the \ac{mocap} actor through a VR headset. TRUMANS \cite{jiang2024scaling}, a recent dataset, marks an advancement by capturing a diverse range of human activities in indoor scenes with dynamically involved objects. However, TRUMANS only provides frame-wise action labels without contextually separable clips, which restricts its applicability in training text-guided motion synthesis models.
To address the shortcomings of existing datasets, we introduce a large-scale dataset by realizing real-world motion capture through a VR-assisted setup. The improved efficiency for deployment enhances the dataset diversity encompassing locomotion in various cluttered scenes, grasping, and \ac{hoi}. Our dataset is uniquely annotated with detailed action descriptions in natural language.

\begin{figure*}
    \includegraphics[width=\linewidth]{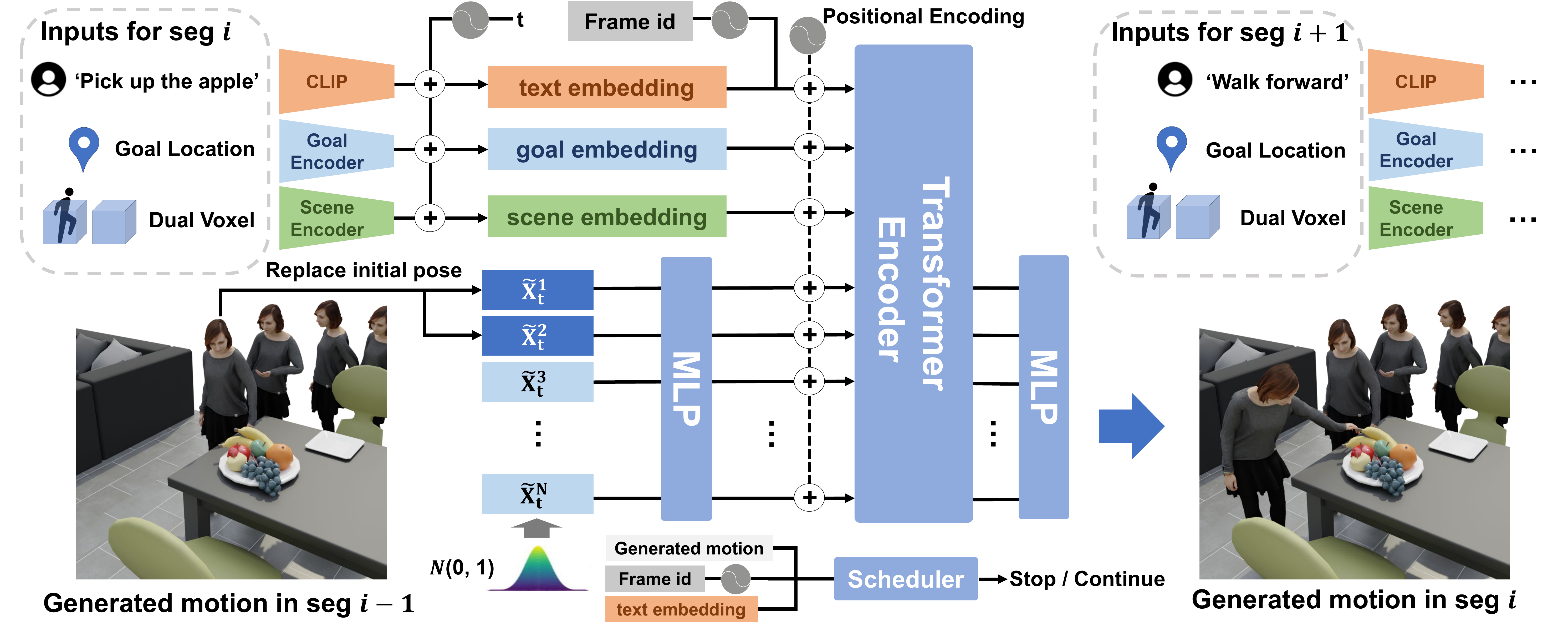}
    \caption{\textbf{Overview of our method}. Our method uses an auto-regressive diffusion model that generates the next motion segment based on existing motions (\cref{sec:method_diffusion}). The 3D environment is captured through a dual voxel scene encoder (\cref{sec:method_scene_embed}). The text instructions are encoded with the time frame to provide precise and time-specific semantic guidance (\cref{sec:method_text_embed}). The goal encoder (\cref{sec:method_goal_embed}) embeds the sub-goal locations for different interaction stages, which are automatically determined by our autonomous scheduler (\cref{sec:method_scheduler}).}
    \label{fig:method}
\end{figure*}

\section{Method}

In this paper, we present a framework for \ac{hsi} synthesis that seamlessly integrates locomotion, hand-reaching motion, and \ac{hoi} into a unified model. Our approach enables users to control a virtual character and execute complex interactions using the goal location and simple text instructions as input, which provide concise commands of ``go somewhere'' and ``do something.'' Our proposed method automatically generates smooth and realistic human motions to navigate the environment and engage in various actions, eliminating the need for manual animation or separate models for each type of action.

\subsection{Data Representation}

Formally, our objective is to synthesize a human motion sequence $\mathcal{M}$ of length $L$, given a 3D scene $\mathcal{S}$, a textual instruction $\mathcal{V}$, and a goal position $\mathcal{G}$. We also consider the dynamic object $\mathbf{O}$ if relevant. In this case, we produce the corresponding object pose sequence $\{\mathcal{P}_i\}_{i=1}^{L}$, which contains location and rotation. Due to the auto-regressive generation strategy, all the generated human motion and object poses are based on the previous two frames of human motion $\mathcal{M}_{hist}$. 

\paragraph{Human Motion}

We represent human motion $\mathcal{M}$ using the parameterized human model SMPL-X \cite{pavlakos2019expressive}. The motion is initially generated as body joints locations $\{\mathcal{X}_i\}_{i=1}^{L}$, where $\mathcal{X}_i \in \mathbb{R}^{J \times 3}$ represents the 3D positions of $J=28$ selected joints. These joints include 22 from the body, plus an additional two each for the left hand, right hand, and head, to better capture the rotations at the leaf nodes. The joints are fitted to the SMPL-X pose parameters $\theta$, global orientation $\phi$, relaxed hand poses $h$, and root translation $r$, resulting in the posed human mesh.

\paragraph{Conditions}

The 3D scene is represented using a voxel grid $\mathcal{S} \in \{0, 1\}^{N_x \times N_y \times N_z}$, where 1 indicates that the position is occupied by scene objects or unreachable. When the scene includes dynamic objects, their representations are additionally captured by sampling 256 points from the object's surface along with their corresponding normal vectors facing outwards, denoted as $O\in \mathbb{R}^{256\times6}$. 

The goal location is represented as a 3D location $\mathcal{G} \in \mathbb{R}^3$ specified by the user, where the interpretation of $\mathcal{G}$ differs for locomotion tasks and \ac{hoi}, discussed in \cref{sec:method_goal_embed}.

\subsection{Motion Diffusion Module}\label{sec:method_diffusion}

The motion diffusion module is responsible for synthesizing human motions based on given text instructions, scene information, and goals. This module leverages the power of diffusion models, specifically the Denoising Diffusion Probabilistic Models (DDPM) \cite{ho2020denoising}, to generate realistic and coherent human motion sequences. We employ an auto-regressive generation strategy, recently gaining popularity in motion synthesis \cite{chen2024taming}. This strategy allows us to generate motion sequences of arbitrary length by recursively generating motion segments $ X = \{\mathcal{X}_i\}_{i=1}^{W}$ of fixed length $W$.

In the forward diffusion process, random noise is gradually added to the canonicalized input motion data over a series of timesteps. Here, \textit{canonicalized} means the joint locations of the current segment are represented in the coordinates of the character's pelvis bone in the first frame. The reverse diffusion process learns to denoise the corrupted motion data, starting from the pure noise denoted as $\tilde{X}_{T}$ and progressively removing the noise to recover the original motion. 

Following \citet{ho2020denoising}, we train a network $\epsilon_\theta$ to predict the noise $\epsilon$ that was added to the original data at each timestep, based on the noisy motion data $\tilde{X}_{t}$, the corresponding timestep $t$, and the condition terms $\mathcal{C}=\{\mathcal{S}_{emb},\mathcal{V}_{emb},\mathcal{G}_{emb}\}$ introduced in the three sections followed. The learning objective thus follows a simple loss:
\begin{equation}
    \mathcal{L}=\mathbb{E}_{\tilde{X}_t\sim q(\tilde{X}_t|\mathcal{C}),t\sim U(1,T)}||{\epsilon - \epsilon_\theta(\tilde{X}_t, t, \mathcal{C})}||_{2}^2.
\end{equation}

\begin{figure*}
  \includegraphics[width=\linewidth]{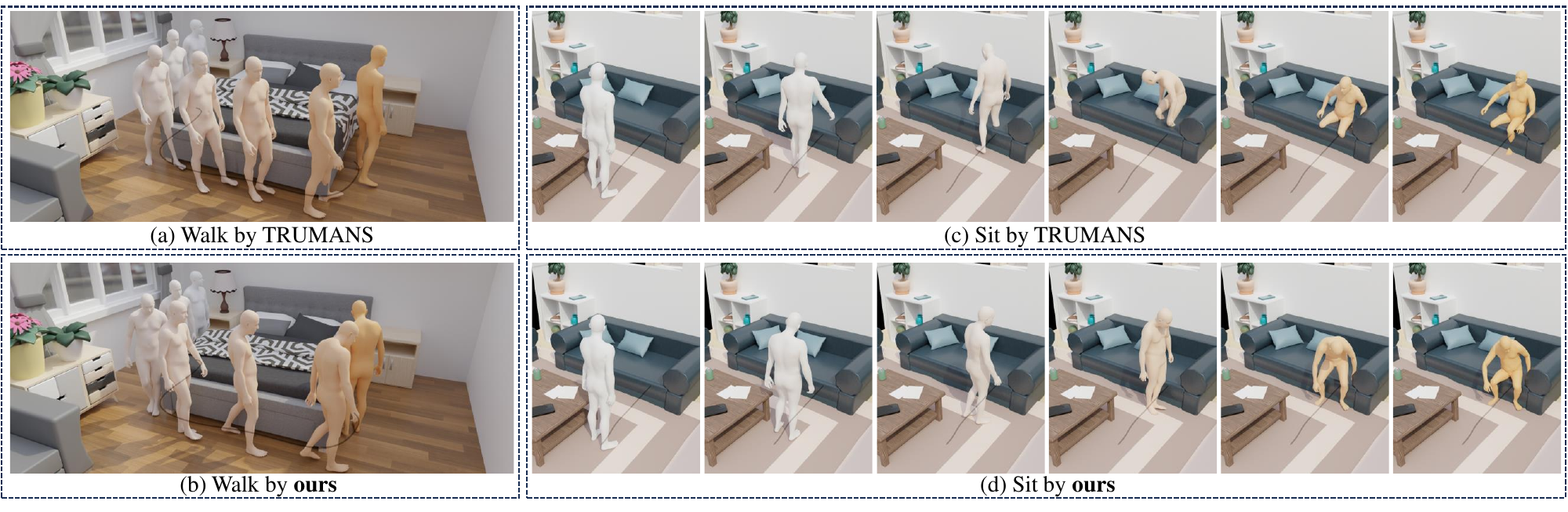}
  \caption{\textbf{Comparison results.} We qualitatively compare our method with TRUMANS \cite{jiang2024scaling}. The left side shows the locomotion along a trajectory, and the right side shows the interaction of sitting on the sofa. Our method generates characters that actively avoid penetrating the scene and exhibit natural cues of scene awareness. For more qualitative results, we refer readers to the supplementary video.}
  \label{fig:cmp}
\end{figure*}

\subsection{Dual Voxel Scene Encoder}\label{sec:method_scene_embed}

To achieve responsive motions within the 3D scene, including collision avoidance and natural interactions (\eg, sitting and lying), the character must be aware of not only the immediate local scene but also the forthcoming environment with which it will engage in the near future. To address this challenge, we introduce a dual voxel scene representation.

Our approach builds on the work presented in TRUMANS \cite{jiang2024scaling}, utilizing a 3D occupancy voxel grid to encapsulate local scene information. The primary mechanism involves constructing a $32\times32\times32$ grid centered around the location of interest, covering an area where each side spans 1.2 meters. Each cell within this grid is then queried in the scene mesh for occupancy, resulting in a binary 3D array where 1 indicates an occupied cell and 0 indicates an unoccupied one.

Given the dynamic nature of human motion and interaction, our method does not rely on a static scene representation. Instead, it employs a dual voxel system designed to capture both the current and imminent scene contexts that influence the character's movement:
\begin{itemize}[nolistsep,noitemsep,leftmargin=*,topsep=0pt]
    \item \textit{Current Scene Voxel}: This voxel grid is centered on the pelvis location of the character at the first frame of the current motion segment, aligning with the character's orientation at that frame.
    \item \textit{Predictive Scene Voxel}: The position of this voxel depends on the stage of the motion segment. For locomotion, the Predictive Scene Voxel is placed 0.8 meters away in the direction of the goal condition. This voxel is directly aligned with the object's location for interactions with specific scene objects, ensuring that the character's movements are both anticipatory and contextually appropriate. Predictive Scene Voxel does not apply to interactions with hand-held small objects and is copied from Current Scene Voxel. The orientation of the voxel is aligned in a way akin to the Current Scene Voxel.
\end{itemize}

To extract meaningful features $\mathcal{S}_{emb}$ from the 3D voxel data, we employ a \ac{vit} architecture \cite{dosovitskiy2020vit}. The vertical dimension of the voxel grid (height) is considered the channel dimension (akin to color channels in images), and the remaining two dimensions (width and depth) serve as the spatial dimensions (similar to width and height in images). 

\subsection{Frame-embedded Text Encoder}\label{sec:method_text_embed}

Unlike traditional text-to-motion methods that directly generate an entire sequence of motion from a sentence, our approach auto-regressively generates motion for the next short period without pre-defined timing. This challenges seamlessly linking multiple segments into a full and semantically correct motion based on the text instruction. For example, a target motion clip labeled ``read a book'' includes the full motion sequence of first turning the page and then reading. To address this, we propose embedding the frame number of action execution into the text during the training and sampling phases.

More specifically, the frame number to be embedded is the number of the first frame in the currently generated motion segment, referenced from the beginning of the original semantically meaningful motion clip. We utilize a sinusoidal positional encoder to convert the frame number from an integer to a 512-dimensional vector, forming the frame embedding. The frame embedding is then added to the embedding of the input textual guidance $\mathcal{V}_{emb}\in\mathbb{R}^{512}$ from the CLIP encoder \cite{radford2021learning}. The summation of these two vectors provides the text conditioning token for the motion synthesis model.

By integrating the frame embedding into the text encoding process, our method learns the temporal patterns of semantic motion, ensuring the generation of coherent and contextually accurate motion segments and effectively linking them into a seamless sequence.

\subsection{Stage-specific Goal}\label{sec:method_goal_embed}

The stage-specific goal represents one location $\mathcal{G} \in \mathbb{R}^3$ in the scene as the condition for the current segment. To fit into the auto-regressive generation pipeline, $\mathcal{G}$ is represented in the character's pelvis coordinate of the first frame. $\mathcal{G}$ goes through MLPs to obtain the goal embedding $\mathcal{G}_{emb}$ as a condition for the motion diffusion module. The goal here is not forced to reach, unlike TRUMANS, which controls character movement by fixing pelvis or hand joints at specific frames. Our approach leaves more space for the generative model to decide how to reach the goal autonomously.

For locomotion, the goal term is a two-dimensional vector representing the walking direction of the current segment multiplied by a user-specified speed. The direction is derived by sampling a walkable sub-goal near the segment's starting point, which is close to the direction towards the intended goal location. Walkability is determined by checking if the line segment to the sub-goal intersects any unwalkable areas. We neglect the vertical component of $\mathcal{G}$, focusing on horizontal movement. The method can also accommodate predefined trajectories, where the trajectory ahead determines the direction.

In hand grasping interactions, like ``pick up the apple,'' $\mathcal{G}$ corresponds to the position of the index finger. This goal term is similarly applied to put-down actions, setting the goal to the finger's release position. In both settings, the goal term guides the hand towards the target position, and the object is attached to the hand/released at the frame in which the index finger is closest to the target. We apply guidance to the last ten diffusion steps to avoid hand-object penetration, inspired by SceneDiffuser \cite{huang2023sceneDiffuser}. When a penetration occurs, we first identify the closest surface points for each penetrated hand joint. Next, we calculate the average direction of the normal of these surface points as the out-moving direction. To resolve the penetration, we shift the predicted denoised hand joints along this out-moving direction.

For scene-level interactions with specific 3D objects, such as ``sit on the sofa,'' the character receives $\mathcal{G}$ as a precise location indicating the pelvis position for the interaction, like the place to sit. The goal embedding is set to zero for interactions involving small objects, as the body location is not specified during the interaction.

\subsection{Autonomous Scheduler}\label{sec:method_scheduler}

An applicable solution to complex instruction is to leverage foundation models to decompose the instruction into manageable stages, including locomotion and interaction. The autonomous scheduler is trained to determine the optimal stopping point for each stage of the interaction process. The autonomous scheduler decides whether to conclude the current stage and move on to the next based on the latest motion segment, the current frame number, and the given language instruction. Its architecture closely resembles that of the motion diffusion module, leveraging a Transformer Encoder. The frame number and language instruction are embedded in the same way as the Frame-embedded Text Encoder (\cref{sec:method_text_embed}), which occupies one input token in the Transformer Encoder. Each frame of the encoded motion is represented by a single token that occupies the rest of the tokens. The autonomous scheduler predicts a value between 0 and 1, indicating the likelihood of the current stage concluding at the present motion segment. 

\section{LINGO dataset}

We propose LINGO, a large-scale \ac{mocap} dataset featuring \acp{hsi}, including locomotion, grasping, and \ac{hoi}. The dataset includes various indoor environments, such as bedrooms, dining rooms, offices, and shops, with 3D models sourced from artists on online marketplaces. Skilled actors with motion capture experience performed a variety of interactions, following detailed instructions for common indoor activities like drinking water, playing instruments, using electronic devices, and engaging in physical activities. The LINGO dataset and the code will be made publicly available for research purposes.

\subsection{VR-assisted \acs{mocap}}

\begin{figure}[t!]
    \centering
    \begin{subfigure}{\linewidth}
        \centering
        \includegraphics[width=\linewidth]{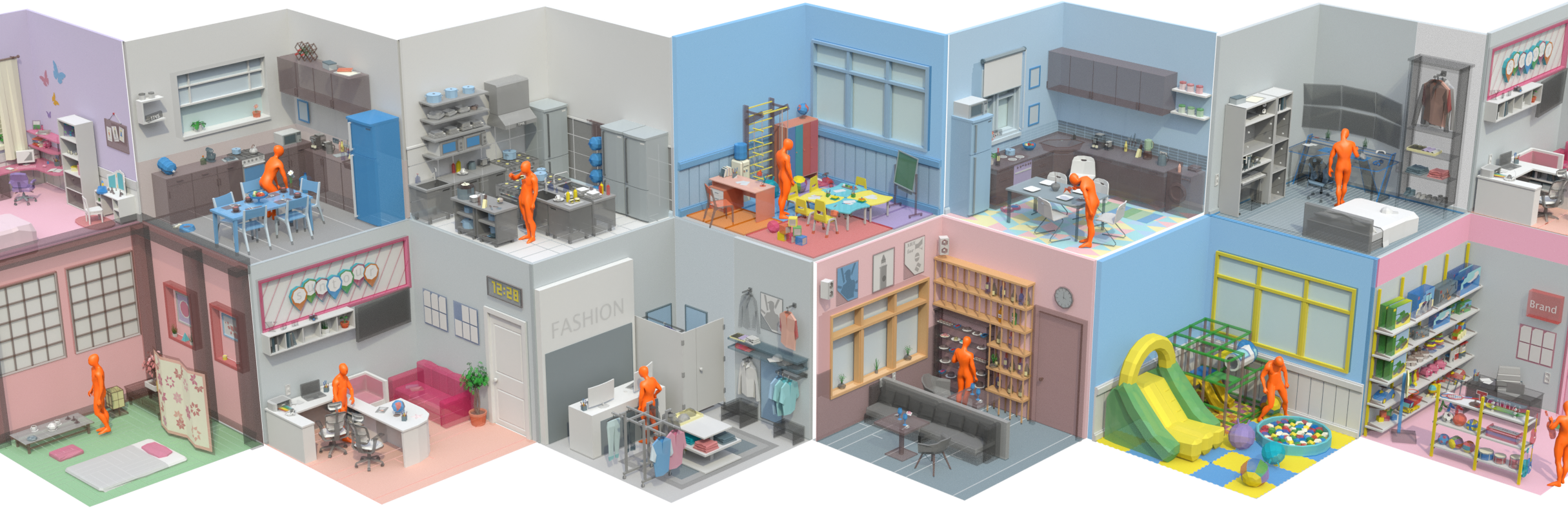}
        \caption{Selected frames from the LINGO dataset}
        \label{fig:subfig1}
    \end{subfigure}
    \begin{subfigure}{0.632\linewidth}
        \centering
        \includegraphics[width=\linewidth]{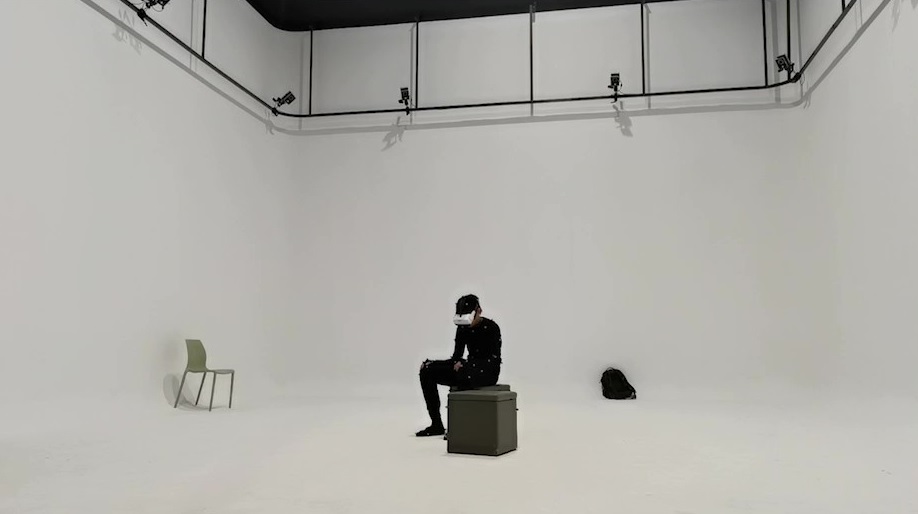}
        \caption{\acs{mocap} setup}
        \label{fig:subfig2}
    \end{subfigure}
    \hfill
    \begin{subfigure}{0.356\linewidth}
        \centering
        \includegraphics[width=\linewidth]{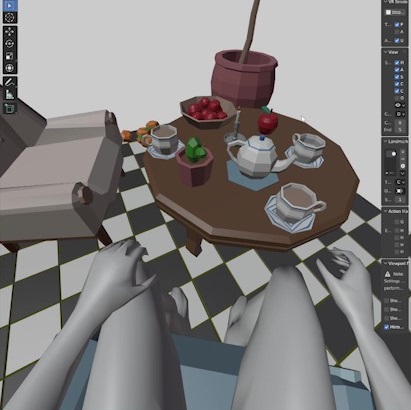}
        \caption{View in VR}
        \label{fig:subfig3}
    \end{subfigure}
    \caption{\textbf{LINGO dataset.} We show some selected frames and the setup of the VR-assisted \acs{mocap}.}
    \label{fig:lingo}
\end{figure}

Constructing human-scene interaction datasets in the physical world presents two significant challenges: the resource-intensive process of capturing a wide variety of scenes and the difficulty in obtaining accurate pose data to effectively incorporate dynamic objects. To address the challenge, LINGO leverages a ``synthetic vision'' projected on a VR headset worn by the motion actor, as \cref{fig:lingo} shows. Building on the concept introduced by the CIRCLE dataset \cite{araujo2023circle}, LINGO goes a step further by incorporating interactive objects, both static and dynamic, into captured scenes.

For static objects like chairs and beds, a sittable object with the same height as the virtual object is placed in the physical space, matching the scene in the VR environment. This enables the actor to physically interact with the object while maintaining the illusion of the virtual environment. Capturing dynamic object interactions poses an additional challenge, as the objects must move with the actor's hand during grasping and manipulation. LINGO incorporates a human assistant during the interaction data recording process to overcome this. The assistant precisely marks the frame when the actor grasps the object, at which point the object becomes bound to the actor's hand and follows their movements seamlessly. When the actor releases the object, the assistant marks the moment, unbinding the object from the actor's hand.

We utilize the advanced VICON \ac{mocap} system, which results in higher motion data quality than multi-view cameras \cite{hassan2019resolving} or IMU-based sensors \cite{vonMarcard2018} commonly used in traditional methods. VR also helps minimize occlusions by avoiding large furniture within the VICON space. 

\subsection{Statistics and Details}

\begin{figure*}
    \includegraphics[width=0.9\linewidth]{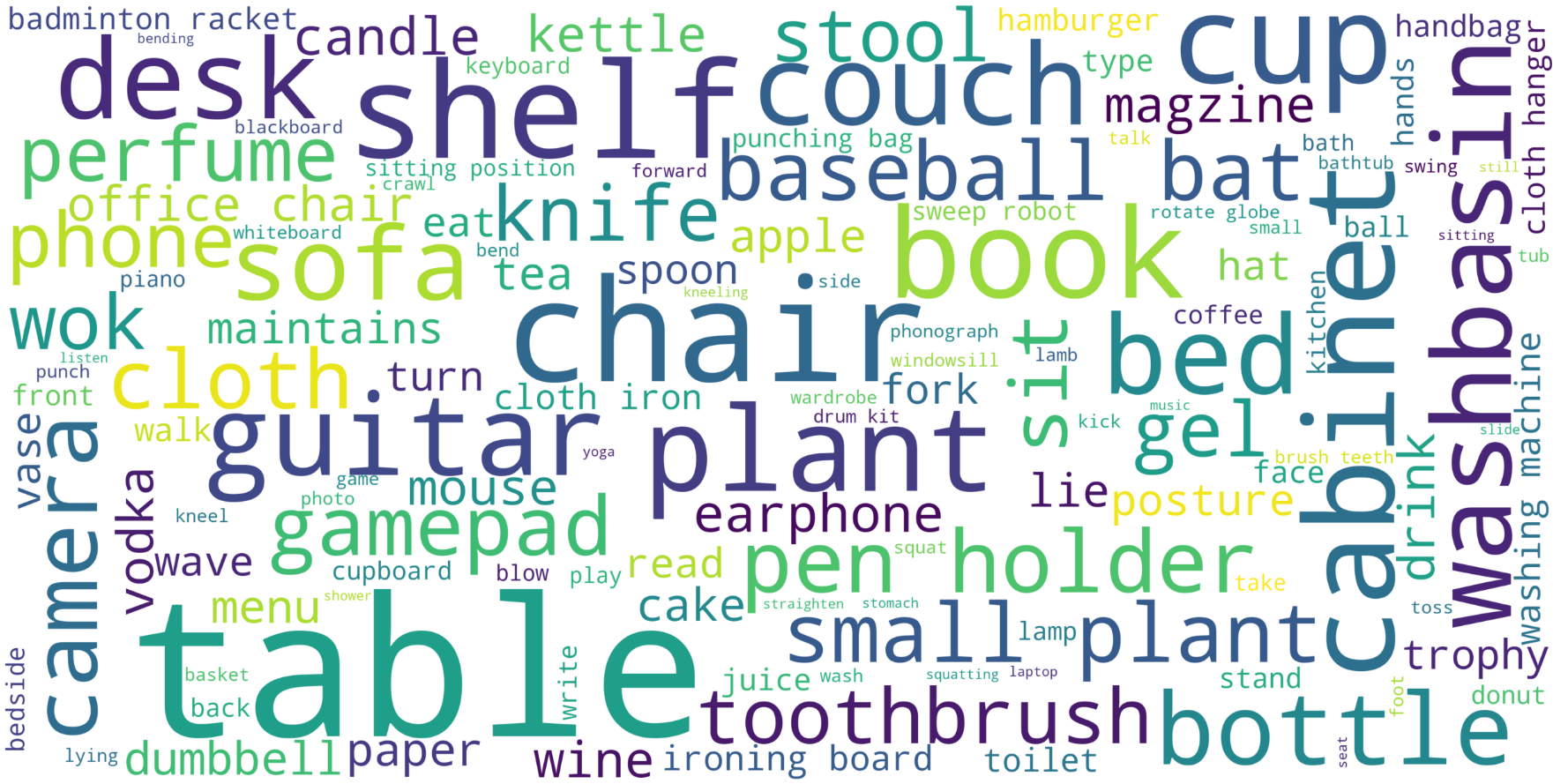}
    \caption{\textbf{Word cloud built from the language annotations in the LINGO dataset.}}
    \label{fig:wordcloud}
\end{figure*}

LINGO features 16 hours of motion sequences captured in 120 unique indoor scenes. The dataset covers a wide spectrum of 40 motion types, each accompanied by precise language descriptions for clarity and usability, summarized in \cref{fig:motion_cnt,fig:wordcloud}. The language descriptions are further augmented for generalizability. The motion sequences are organized into 4 to 6 long-term segments per scene, each spanning 1 to 3 minutes. The human motion is captured at 30 FPS, leveraging a 53-point FrontWaist \ac{mocap} suit. The LINGO-train and LINGO-eval sets are split by the scenes, ensuring a 4:1 proportion for each room type without overlap. Accompanying the interactions, the LINGO dataset incorporates 20 types of objects commonly used in everyday life.

To represent human subjects, LINGO employs the SMPL-X model \cite{pavlakos2019expressive}. The dataset utilizes the MANO hand model with 12 PCA components for efficient hand representation. Facial expressions are not captured in this dataset. For more detailed information regarding the dataset, such as the specific interaction types, object categories, and scene descriptions, the readers are encouraged to refer to the supplementary material.

\begin{table*}[ht!]
    \caption{\textbf{Quantitative results of interactive motion synthesis.} The instructions involve performing interaction with an object in the scene.}
    \label{tab:semantic}
    \begin{tabular}{lcccccc}
        \hline
        Interactive motion   & FID$\downarrow$             & Diversity$\rightarrow$    & Multi-modality$\rightarrow$ & Precision$\uparrow$       & Recall$\uparrow$          & F1 score$\uparrow$        \\ \hline
        Real motions         & -                           & 6.379                     & 1.119                       & 0.888                     & 0.907                     & 0.907                     \\
        TRUMANS              & $2.438{\pm.041}$            & $6.182{\pm.078}$          & $3.353{\pm.077}$            & $0.628{\pm.004}$          & $0.557{\pm.004}$          & $0.552{\pm.004}$          \\
        \textbf{Ours}        & \textbf{2.048${\pm.058}$}   & $6.220{\pm.076}$          & \textbf{2.919${\pm.081}$}   & \textbf{0.695${\pm.004}$} & \textbf{0.629${\pm.004}$} & \textbf{0.622${\pm.004}$} \\
        w/o frame embedding  & 2.368${\pm.070}$            & \textbf{6.300${\pm.062}$} & $3.600{\pm.072}$            & $0.615{\pm.005}$          & $0.553{\pm.005}$          & $0.540{\pm.006}$          \\ \hline
    \end{tabular}
\end{table*}

\begin{table}[ht!]
    \caption{\textbf{Quantitative results of locomotion}, where the character walks from one place to another in cluttered scenes.}
    \label{tab:locomotion}
    \resizebox{\linewidth}{!}{
        \begin{tabular}{lcccc}
            \hline
            Locomotion      & Pene$_{\%scene}$$\downarrow$  & Pene$_{mean}$$\downarrow$  & Pene$_{max}$$\downarrow$  & FS$\downarrow$            \\ \hline
            TRUMANS         & $0.048{\pm.006}$              & $1.011{\pm.012}$           & $7.441{\pm.383}$          & $0.472{\pm.013}$          \\
            \textbf{Ours}   & \textbf{0.038${\pm.001}$}     & \textbf{0.402${\pm.004}$}  & \textbf{0.948${\pm.065}$} & \textbf{0.432${\pm.004}$} \\
            flattened voxel & $0.045{\pm.002}$              & $0.587{\pm.005}$           & $3.373{\pm.117}$          & $0.470{\pm.030}$          \\ \hline
        \end{tabular}%
    }%
\end{table}

\begin{table}[ht!]
    \caption{\textbf{Quantitative results of object reaching}, where the character is instructed to walk toward and reach for an object.}        \label{tab:reaching}
    \resizebox{\linewidth}{!}{
        \begin{tabular}{llll}
            \hline
            Object reaching     & Error dist.$\downarrow$   & Pene$_{\%scene}$$\downarrow$ & Time used$\downarrow$     \\ \hline
            GOAL                & $0.156{\pm.028}$          & $0.057{\pm.004}$             & $4.880{\pm.356}$          \\
            \textbf{Ours}       & \textbf{0.061${\pm.004}$} & \textbf{0.045${\pm.003}$}    & \textbf{3.073${\pm.351}$} \\
            w/o dual voxel      & $0.111{\pm.013}$          & $0.057{\pm.008}$             & $3.147{\pm.360}$          \\
            w/o frame embedding & $0.135{\pm.026}$          & $0.046{\pm.003}$             & $3.980{\pm.242}$          \\ \hline
        \end{tabular}%
    }%
\end{table}

\section{Experiments}

Our experimental evaluation is divided into three primary settings associated with three stages of motion: locomotion, object reaching, and interactive motion. We compare our method with state-of-the-art methods leveraging generative models. All comparison methods are trained in LINGO-train and evaluated in LINGO-eval.

\subsection{Evaluation Settings}

\paragraph{Locomotion}

In the locomotion experiment, we investigate the character's ability to navigate a scene while avoiding collisions and maintaining natural movement. We randomly generate 100 pairs of start and end locations, ensuring that these direct paths intersect with objects in the scene, thereby testing the character's collision avoidance skills in cluttered environments. We compare our method with TRUMANS \cite{jiang2024scaling} and include an ablation study to evaluate the significance of dual voxel scene representation. In this ablation case, we flatten the 3D scene voxel along the vertical axis, forming a walkable map where 0 indicates a walkable area and 1 indicates an occupied location. For evaluation, we measure scene penetration using metrics adapted from DIMOS \cite{zhao2023synthesizing}, which include the ratio of body vertices that penetrate scene objects (Pene$_{\%scene}$), the average penetration distance throughout the synthesized sequence (Pene$_{mean}$), and the maximum penetration distance in any frame during the sequence (Pene$_{max}$). Furthermore, we assess foot sliding using the metric from \cite{he2022nemf}. 

\paragraph{Object Reaching}

We evaluate the character's ability to reach for an object while being aware of the surrounding environment. We randomly select 100 pairs of starting positions for the character and goal positions for the object. The evaluation focuses on both the quality of the reaching motion and collision with scenes. We compare the results with GOAL \cite{taheri2021Goal} as a baseline. We include two ablation studies where the frame embedding or dual voxel is removed. Metrics used include the error distance between the intended goal position and the actual hand position after reaching the object, the same scene penetration metrics as in the locomotion setting (Pene$_{\%scene}$), and the time to complete the reaching task (no more than 20 cm between the hand and the object) measured in seconds.

\paragraph{Interactive Motion}

In this setting, we generate motions that involve interaction with objects from the LINGO dataset. For small objects like bottles and gamepads, we assume that the objects are already grasped and focus on the semantic quality of the motion. For larger objects such as sofas and beds, the character first walks towards the object and then interacts by sitting or lying down. We compare our method with TRUMANS, using several metrics following \cite{tevet2022human}, including \ac{fid} to measure motion realism, Diversity and Multi-modality to evaluate the range of different motions produced, Precision to measure how closely the generated motions match the reference, Recall to determine the coverage of the reference motions by the generated ones, and the F1 score to provide a balanced measure. We modify the action encoder of TRUMANS to use the same language encoder as our method.

\subsection{Results}

\begin{figure*}[t!]
    \centering
    \begin{subfigure}{0.9\linewidth}
        \centering
        \includegraphics[width=\linewidth]{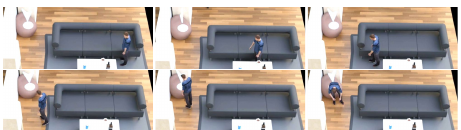}
        \caption{}
        \label{fig:cmp1}
    \end{subfigure}
    \begin{subfigure}{0.9\linewidth}
        \centering
        \includegraphics[width=\linewidth]{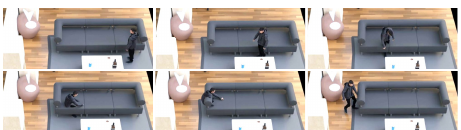}
        \caption{}
        \label{fig:cmp2}
    \end{subfigure}
    \begin{subfigure}{0.9\linewidth}
        \centering
        \includegraphics[width=\linewidth]{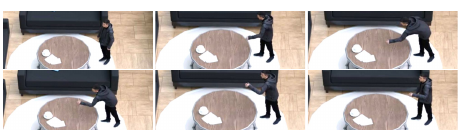}
        \caption{}
        \label{fig:cmp3}
    \end{subfigure}
    \begin{subfigure}{0.9\linewidth}
        \centering
        \includegraphics[width=\linewidth]{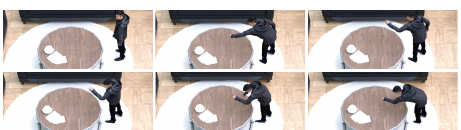}
        \caption{}
        \label{fig:cmp4}
    \end{subfigure}
    \caption{\textbf{Qualitative comparison.} We compare (a) our method with (b) TRUMANS \cite{jiang2024scaling} on the task of walking to the goal location. It is shown that our method is aware of the surroundings for collision avoidance, while TRUMANS depends on a pre-defined trajectory. We show (c) our method and (d) w/o frame embedder given ``grasp an object'' instruction.  The synthesized motion without a frame embedder is disordered and tends to repeat.}
    \label{fig:compare}
\end{figure*}

\cref{tab:semantic} presents the quantitative results of semantic motion synthesis, demonstrating that our method achieves the highest scores on all metrics except Diversity. The ablation study, which removes the frame embedding, results in higher Diversity because the synthesized motions become disordered, often repeating the same actions without maintaining a coherent flow. It lacks awareness of the overall motion sequence, which can be observed in \cref{fig:cmp3,fig:cmp4}. All other metrics consistently show that the motion synthesized by our method is of high quality and maintains coherence with the given semantics. This indicates that our approach effectively balances realism and semantic alignment, producing natural and contextually correct motions.

\cref{tab:locomotion} shows the quantitative results for the locomotion task. Our method outperforms TRUMANS in all evaluated metrics. The scene penetration from our method is significantly lower, highlighting its superior capability to autonomously plan and adjust trajectories to avoid collisions rather than following a predefined path. This autonomous decision-making is crucial for navigating cluttered scenes, where subtle body adjustments are necessary to avoid obstacles, demonstrated in \cref{fig:cmp,fig:cmp1,fig:cmp2}. Additionally, the foot sliding metric for our method is also low, indicating high-quality motion with stable and realistic movement. From the ablations where the dual voxel representation is replaced with a flattened walkable map, the higher penetration and foot sliding underscores the effectiveness of our dual voxel scene representation in providing comprehensive 3D information about the surroundings, enabling more precise and realistic motion adjustments.

\cref{tab:reaching} confirms the efficiency and robustness of our method in synthesizing realistic hand-oriented motions. Specifically, our method achieves the best scores in terms of low reaching error, low penetration with the scene, and short time used for reaching the goal. The ablations further validate the efficacy of our dual voxel scene encoder and the frame embedding. \cref{tab:locomotion,tab:reaching} jointly demonstrate that our method can generalize to unseen environments. \cref{fig:cmp3,fig:cmp4} shows that the frame embedding helps to generate semantically coherent motions.

\section{Conclusion}

\paragraph{Limitations} (i) Our approach concentrates on body-level motions, ignoring the intricate details of hand-level manipulation and facial expressions. (ii) Although quantitative results indicate that our method achieves superior scene awareness compared to existing techniques, it does not guarantee a perfect physical plausibility of the generated motions. (iii) We have not researched the generalizability of unseen interaction types in this work. 

In summary, this work introduces a comprehensive generative framework that autonomously synthesizes multi-stage, scene-aware human motions directly from text instructions and goal locations. We present an approach that seamlessly integrates human motions with 3D scenes with a novel 3D scene representation and a joint time frame and language embedding. We also contribute a detailed, language-annotated \ac{mocap} dataset, providing a valuable resource for future research in human motion synthesis.

\begin{acks}

The authors would like to thank NVIDIA for generously providing the necessary GPUs and hardware support. This work is supported in part by the National Science and Technology Major Project (2022ZD0114900), an NSFC fund (62376009), and the Beijing Nova Program.



\end{acks}

\bibliographystyle{ACM-Reference-Format}
\bibliography{reference_header,reference}


\begin{thebibliography}{49}


\ifx \showCODEN    \undefined \def \showCODEN     #1{\unskip}     \fi
\ifx \showDOI      \undefined \def \showDOI       #1{#1}\fi
\ifx \showISBNx    \undefined \def \showISBNx     #1{\unskip}     \fi
\ifx \showISBNxiii \undefined \def \showISBNxiii  #1{\unskip}     \fi
\ifx \showISSN     \undefined \def \showISSN      #1{\unskip}     \fi
\ifx \showLCCN     \undefined \def \showLCCN      #1{\unskip}     \fi
\ifx \shownote     \undefined \def \shownote      #1{#1}          \fi
\ifx \showarticletitle \undefined \def \showarticletitle #1{#1}   \fi
\ifx \showURL      \undefined \def \showURL       {\relax}        \fi
\providecommand\bibfield[2]{#2}
\providecommand\bibinfo[2]{#2}
\providecommand\natexlab[1]{#1}
\providecommand\showeprint[2][]{arXiv:#2}

\bibitem[Ara{\'u}jo et~al\mbox{.}(2023)]%
        {araujo2023circle}
\bibfield{author}{\bibinfo{person}{Joao~Pedro Ara{\'u}jo}, \bibinfo{person}{Jiaman Li}, \bibinfo{person}{Karthik Vetrivel}, \bibinfo{person}{Rishi Agarwal}, \bibinfo{person}{Jiajun Wu}, \bibinfo{person}{Deepak Gopinath}, \bibinfo{person}{Alexander~William Clegg}, {and} \bibinfo{person}{Karen Liu}.} \bibinfo{year}{2023}\natexlab{}.
\newblock \showarticletitle{CIRCLE: Capture In Rich Contextual Environments}. In \bibinfo{booktitle}{\emph{Conference on Computer Vision and Pattern Recognition (CVPR)}}.
\newblock


\bibitem[Bhatnagar et~al\mbox{.}(2022)]%
        {bhatnagar22behave}
\bibfield{author}{\bibinfo{person}{Bharat~Lal Bhatnagar}, \bibinfo{person}{Xianghui Xie}, \bibinfo{person}{Ilya~A. Petrov}, \bibinfo{person}{Cristian Sminchisescu}, \bibinfo{person}{Christian Theobalt}, {and} \bibinfo{person}{Gerard Pons-Moll}.} \bibinfo{year}{2022}\natexlab{}.
\newblock \showarticletitle{BEHAVE: Dataset and Method for Tracking Human Object Interactions}. In \bibinfo{booktitle}{\emph{Conference on Computer Vision and Pattern Recognition (CVPR)}}.
\newblock


\bibitem[Chen et~al\mbox{.}(2024)]%
        {chen2024taming}
\bibfield{author}{\bibinfo{person}{Rui Chen}, \bibinfo{person}{Mingyi Shi}, \bibinfo{person}{Shaoli Huang}, \bibinfo{person}{Ping Tan}, \bibinfo{person}{Taku Komura}, {and} \bibinfo{person}{Xuelin Chen}.} \bibinfo{year}{2024}\natexlab{}.
\newblock \showarticletitle{Taming Diffusion Probabilistic Models for Character Control}. In \bibinfo{booktitle}{\emph{SIGGRAPH Conference Papers}}.
\newblock


\bibitem[Coros et~al\mbox{.}(2010)]%
        {coros2010generalized}
\bibfield{author}{\bibinfo{person}{Stelian Coros}, \bibinfo{person}{Philippe Beaudoin}, {and} \bibinfo{person}{Michiel Van~de Panne}.} \bibinfo{year}{2010}\natexlab{}.
\newblock \showarticletitle{Generalized biped walking control}.
\newblock \bibinfo{journal}{\emph{ACM Transactions on Graphics (TOG)}} \bibinfo{volume}{29}, \bibinfo{number}{4} (\bibinfo{year}{2010}), \bibinfo{pages}{1--9}.
\newblock


\bibitem[Cui et~al\mbox{.}(2024)]%
        {cui2024anyskill}
\bibfield{author}{\bibinfo{person}{Jieming Cui}, \bibinfo{person}{Tengyu Liu}, \bibinfo{person}{Nian Liu}, \bibinfo{person}{Yaodong Yang}, \bibinfo{person}{Yixin Zhu}, {and} \bibinfo{person}{Siyuan Huang}.} \bibinfo{year}{2024}\natexlab{}.
\newblock \showarticletitle{Anyskill: Learning Open-Vocabulary Physical Skill for Interactive Agents}. In \bibinfo{booktitle}{\emph{Conference on Computer Vision and Pattern Recognition (CVPR)}}.
\newblock


\bibitem[Dosovitskiy et~al\mbox{.}(2021)]%
        {dosovitskiy2020vit}
\bibfield{author}{\bibinfo{person}{Alexey Dosovitskiy}, \bibinfo{person}{Lucas Beyer}, \bibinfo{person}{Alexander Kolesnikov}, \bibinfo{person}{Dirk Weissenborn}, \bibinfo{person}{Xiaohua Zhai}, \bibinfo{person}{Thomas Unterthiner}, \bibinfo{person}{Mostafa Dehghani}, \bibinfo{person}{Matthias Minderer}, \bibinfo{person}{Georg Heigold}, \bibinfo{person}{Sylvain Gelly}, \bibinfo{person}{Jakob Uszkoreit}, {and} \bibinfo{person}{Neil Houlsby}.} \bibinfo{year}{2021}\natexlab{}.
\newblock \showarticletitle{An Image is Worth 16x16 Words: Transformers for Image Recognition at Scale}. In \bibinfo{booktitle}{\emph{International Conference on Learning Representations (ICLR)}}.
\newblock


\bibitem[Fan et~al\mbox{.}(2023)]%
        {fan2023arctic}
\bibfield{author}{\bibinfo{person}{Zicong Fan}, \bibinfo{person}{Omid Taheri}, \bibinfo{person}{Dimitrios Tzionas}, \bibinfo{person}{Muhammed Kocabas}, \bibinfo{person}{Manuel Kaufmann}, \bibinfo{person}{Michael~J. Black}, {and} \bibinfo{person}{Otmar Hilliges}.} \bibinfo{year}{2023}\natexlab{}.
\newblock \showarticletitle{{ARCTIC}: A Dataset for Dexterous Bimanual Hand-Object Manipulation}. In \bibinfo{booktitle}{\emph{Conference on Computer Vision and Pattern Recognition (CVPR)}}.
\newblock


\bibitem[Ghosh et~al\mbox{.}(2023)]%
        {ghosh2022imos}
\bibfield{author}{\bibinfo{person}{Anindita Ghosh}, \bibinfo{person}{Rishabh Dabral}, \bibinfo{person}{Vladislav Golyanik}, \bibinfo{person}{Christian Theobalt}, {and} \bibinfo{person}{Philipp Slusallek}.} \bibinfo{year}{2023}\natexlab{}.
\newblock \showarticletitle{IMoS: Intent-Driven Full-Body Motion Synthesis for Human-Object Interactions}. In \bibinfo{booktitle}{\emph{Eurographics}}.
\newblock


\bibitem[Guzov et~al\mbox{.}(2023)]%
        {guzov23ireplica}
\bibfield{author}{\bibinfo{person}{Vladimir Guzov}, \bibinfo{person}{Julian Chibane}, \bibinfo{person}{Riccardo Marin}, \bibinfo{person}{Yannan He}, \bibinfo{person}{Yunus Saracoglu}, \bibinfo{person}{Torsten Sattler}, {and} \bibinfo{person}{Gerard Pons-Moll}.} \bibinfo{year}{2023}\natexlab{}.
\newblock \showarticletitle{Interaction Replica: Tracking human–object interaction and scene changes from human motion}. In \bibinfo{booktitle}{\emph{International Conference on 3D Vision (3DV)}}.
\newblock


\bibitem[Hassan et~al\mbox{.}(2021)]%
        {hassan2021stochastic}
\bibfield{author}{\bibinfo{person}{Mohamed Hassan}, \bibinfo{person}{Duygu Ceylan}, \bibinfo{person}{Ruben Villegas}, \bibinfo{person}{Jun Saito}, \bibinfo{person}{Jimei Yang}, \bibinfo{person}{Yi Zhou}, {and} \bibinfo{person}{Michael Black}.} \bibinfo{year}{2021}\natexlab{}.
\newblock \showarticletitle{Stochastic Scene-Aware Motion Prediction}. In \bibinfo{booktitle}{\emph{International Conference on Computer Vision (ICCV)}}.
\newblock


\bibitem[Hassan et~al\mbox{.}(2019)]%
        {hassan2019resolving}
\bibfield{author}{\bibinfo{person}{Mohamed Hassan}, \bibinfo{person}{Vasileios Choutas}, \bibinfo{person}{Dimitrios Tzionas}, {and} \bibinfo{person}{Michael~J Black}.} \bibinfo{year}{2019}\natexlab{}.
\newblock \showarticletitle{Resolving 3D human pose ambiguities with 3D scene constraints}. In \bibinfo{booktitle}{\emph{International Conference on Computer Vision (ICCV)}}.
\newblock


\bibitem[Hassan et~al\mbox{.}(2023)]%
        {hassan2023synthesizing}
\bibfield{author}{\bibinfo{person}{Mohamed Hassan}, \bibinfo{person}{Yunrong Guo}, \bibinfo{person}{Tingwu Wang}, \bibinfo{person}{Michael Black}, \bibinfo{person}{Sanja Fidler}, {and} \bibinfo{person}{Xue~Bin Peng}.} \bibinfo{year}{2023}\natexlab{}.
\newblock \showarticletitle{Synthesizing Physical Character-Scene Interactions}. In \bibinfo{booktitle}{\emph{SIGGRAPH Conference Papers}}.
\newblock


\bibitem[He et~al\mbox{.}(2022)]%
        {he2022nemf}
\bibfield{author}{\bibinfo{person}{Chengan He}, \bibinfo{person}{Jun Saito}, \bibinfo{person}{James Zachary}, \bibinfo{person}{Holly Rushmeier}, {and} \bibinfo{person}{Yi Zhou}.} \bibinfo{year}{2022}\natexlab{}.
\newblock \showarticletitle{Nemf: Neural motion fields for kinematic animation}. In \bibinfo{booktitle}{\emph{Advances in Neural Information Processing Systems (NeurIPS)}}.
\newblock


\bibitem[Ho et~al\mbox{.}(2020)]%
        {ho2020denoising}
\bibfield{author}{\bibinfo{person}{Jonathan Ho}, \bibinfo{person}{Ajay Jain}, {and} \bibinfo{person}{Pieter Abbeel}.} \bibinfo{year}{2020}\natexlab{}.
\newblock \showarticletitle{Denoising diffusion probabilistic models}. In \bibinfo{booktitle}{\emph{Advances in Neural Information Processing Systems (NeurIPS)}}.
\newblock


\bibitem[Huang et~al\mbox{.}(2023)]%
        {huang2023sceneDiffuser}
\bibfield{author}{\bibinfo{person}{Siyuan Huang}, \bibinfo{person}{Zan Wang}, \bibinfo{person}{Puhao Li}, \bibinfo{person}{Baoxiong Jia}, \bibinfo{person}{Tengyu Liu}, \bibinfo{person}{Yixin Zhu}, \bibinfo{person}{Wei Liang}, {and} \bibinfo{person}{Song-Chun Zhu}.} \bibinfo{year}{2023}\natexlab{}.
\newblock \showarticletitle{Diffusion-based Generation, Optimization, and Planning in 3D Scenes}. In \bibinfo{booktitle}{\emph{Conference on Computer Vision and Pattern Recognition (CVPR)}}.
\newblock


\bibitem[Jiang et~al\mbox{.}(2023)]%
        {jiang2023full}
\bibfield{author}{\bibinfo{person}{Nan Jiang}, \bibinfo{person}{Tengyu Liu}, \bibinfo{person}{Zhexuan Cao}, \bibinfo{person}{Jieming Cui}, \bibinfo{person}{Zhiyuan Zhang}, \bibinfo{person}{Yixin Chen}, \bibinfo{person}{He Wang}, \bibinfo{person}{Yixin Zhu}, {and} \bibinfo{person}{Siyuan Huang}.} \bibinfo{year}{2023}\natexlab{}.
\newblock \showarticletitle{Full-Body Articulated Human-Object Interaction}. In \bibinfo{booktitle}{\emph{International Conference on Computer Vision (ICCV)}}.
\newblock


\bibitem[Jiang et~al\mbox{.}(2024)]%
        {jiang2024scaling}
\bibfield{author}{\bibinfo{person}{Nan Jiang}, \bibinfo{person}{Zhiyuan Zhang}, \bibinfo{person}{Hongjie Li}, \bibinfo{person}{Xiaoxuan Ma}, \bibinfo{person}{Zan Wang}, \bibinfo{person}{Yixin Chen}, \bibinfo{person}{Tengyu Liu}, \bibinfo{person}{Yixin Zhu}, {and} \bibinfo{person}{Siyuan Huang}.} \bibinfo{year}{2024}\natexlab{}.
\newblock \showarticletitle{Scaling up dynamic human-scene interaction modeling}. In \bibinfo{booktitle}{\emph{Conference on Computer Vision and Pattern Recognition (CVPR)}}.
\newblock


\bibitem[Lee and Joo(2023)]%
        {lee2023locomotion}
\bibfield{author}{\bibinfo{person}{Jiye Lee} {and} \bibinfo{person}{Hanbyul Joo}.} \bibinfo{year}{2023}\natexlab{}.
\newblock \showarticletitle{Locomotion-Action-Manipulation: Synthesizing Human-Scene Interactions in Complex 3D Environments}. In \bibinfo{booktitle}{\emph{International Conference on Computer Vision (ICCV)}}.
\newblock


\bibitem[Li et~al\mbox{.}(2023a)]%
        {li2023controllable}
\bibfield{author}{\bibinfo{person}{Jiaman Li}, \bibinfo{person}{Alexander Clegg}, \bibinfo{person}{Roozbeh Mottaghi}, \bibinfo{person}{Jiajun Wu}, \bibinfo{person}{Xavier Puig}, {and} \bibinfo{person}{C~Karen Liu}.} \bibinfo{year}{2023}\natexlab{a}.
\newblock \showarticletitle{Controllable human-object interaction synthesis}. In \bibinfo{booktitle}{\emph{European Conference on Computer Vision (ECCV)}}.
\newblock


\bibitem[Li et~al\mbox{.}(2023b)]%
        {li2023object}
\bibfield{author}{\bibinfo{person}{Jiaman Li}, \bibinfo{person}{Jiajun Wu}, {and} \bibinfo{person}{C~Karen Liu}.} \bibinfo{year}{2023}\natexlab{b}.
\newblock \showarticletitle{Object motion guided human motion synthesis}.
\newblock \bibinfo{journal}{\emph{ACM Transactions on Graphics (TOG)}} \bibinfo{volume}{42}, \bibinfo{number}{6} (\bibinfo{year}{2023}), \bibinfo{pages}{1--11}.
\newblock


\bibitem[Li et~al\mbox{.}(2019)]%
        {li2019putting}
\bibfield{author}{\bibinfo{person}{Xueting Li}, \bibinfo{person}{Sifei Liu}, \bibinfo{person}{Kihwan Kim}, \bibinfo{person}{Xiaolong Wang}, \bibinfo{person}{Ming-Hsuan Yang}, {and} \bibinfo{person}{Jan Kautz}.} \bibinfo{year}{2019}\natexlab{}.
\newblock \showarticletitle{Putting humans in a scene: Learning affordance in 3d indoor environments}. In \bibinfo{booktitle}{\emph{Conference on Computer Vision and Pattern Recognition (CVPR)}}.
\newblock


\bibitem[Liu and Hodgins(2017)]%
        {liu2017learning}
\bibfield{author}{\bibinfo{person}{Libin Liu} {and} \bibinfo{person}{Jessica Hodgins}.} \bibinfo{year}{2017}\natexlab{}.
\newblock \showarticletitle{Learning to schedule control fragments for physics-based characters using deep q-learning}.
\newblock \bibinfo{journal}{\emph{ACM Transactions on Graphics (TOG)}} \bibinfo{volume}{36}, \bibinfo{number}{3} (\bibinfo{year}{2017}), \bibinfo{pages}{1--14}.
\newblock


\bibitem[Liu and Hodgins(2018)]%
        {liu2018learning}
\bibfield{author}{\bibinfo{person}{Libin Liu} {and} \bibinfo{person}{Jessica Hodgins}.} \bibinfo{year}{2018}\natexlab{}.
\newblock \showarticletitle{Learning basketball dribbling skills using trajectory optimization and deep reinforcement learning}.
\newblock \bibinfo{journal}{\emph{ACM Transactions on Graphics (TOG)}} \bibinfo{volume}{37}, \bibinfo{number}{4} (\bibinfo{year}{2018}), \bibinfo{pages}{1--14}.
\newblock


\bibitem[Merel et~al\mbox{.}(2020)]%
        {merel2020catch}
\bibfield{author}{\bibinfo{person}{Josh Merel}, \bibinfo{person}{Saran Tunyasuvunakool}, \bibinfo{person}{Arun Ahuja}, \bibinfo{person}{Yuval Tassa}, \bibinfo{person}{Leonard Hasenclever}, \bibinfo{person}{Vu Pham}, \bibinfo{person}{Tom Erez}, \bibinfo{person}{Greg Wayne}, {and} \bibinfo{person}{Nicolas Heess}.} \bibinfo{year}{2020}\natexlab{}.
\newblock \showarticletitle{Catch \& carry: reusable neural controllers for vision-guided whole-body tasks}.
\newblock \bibinfo{journal}{\emph{ACM Transactions on Graphics (TOG)}} \bibinfo{volume}{39}, \bibinfo{number}{4} (\bibinfo{year}{2020}), \bibinfo{pages}{39--1}.
\newblock


\bibitem[Monszpart et~al\mbox{.}(2019)]%
        {monszpart2019imapper}
\bibfield{author}{\bibinfo{person}{Aron Monszpart}, \bibinfo{person}{Paul Guerrero}, \bibinfo{person}{Duygu Ceylan}, \bibinfo{person}{Ersin Yumer}, {and} \bibinfo{person}{Niloy~J Mitra}.} \bibinfo{year}{2019}\natexlab{}.
\newblock \showarticletitle{iMapper: interaction-guided scene mapping from monocular videos}.
\newblock \bibinfo{journal}{\emph{ACM Transactions on Graphics (TOG)}} \bibinfo{volume}{38}, \bibinfo{number}{4} (\bibinfo{year}{2019}), \bibinfo{pages}{1--15}.
\newblock


\bibitem[Mordatch et~al\mbox{.}(2012)]%
        {mordatch2012discovery}
\bibfield{author}{\bibinfo{person}{Igor Mordatch}, \bibinfo{person}{Emanuel Todorov}, {and} \bibinfo{person}{Zoran Popovi{\'c}}.} \bibinfo{year}{2012}\natexlab{}.
\newblock \showarticletitle{Discovery of complex behaviors through contact-invariant optimization}.
\newblock \bibinfo{journal}{\emph{ACM Transactions on Graphics (TOG)}} \bibinfo{volume}{31}, \bibinfo{number}{4} (\bibinfo{year}{2012}), \bibinfo{pages}{1--8}.
\newblock


\bibitem[Pavlakos et~al\mbox{.}(2019)]%
        {pavlakos2019expressive}
\bibfield{author}{\bibinfo{person}{Georgios Pavlakos}, \bibinfo{person}{Vasileios Choutas}, \bibinfo{person}{Nima Ghorbani}, \bibinfo{person}{Timo Bolkart}, \bibinfo{person}{Ahmed~AA Osman}, \bibinfo{person}{Dimitrios Tzionas}, {and} \bibinfo{person}{Michael~J Black}.} \bibinfo{year}{2019}\natexlab{}.
\newblock \showarticletitle{Expressive body capture: 3d hands, face, and body from a single image}. In \bibinfo{booktitle}{\emph{Conference on Computer Vision and Pattern Recognition (CVPR)}}.
\newblock


\bibitem[Peng et~al\mbox{.}(2019)]%
        {peng2019mcp}
\bibfield{author}{\bibinfo{person}{Xue~Bin Peng}, \bibinfo{person}{Michael Chang}, \bibinfo{person}{Grace Zhang}, \bibinfo{person}{Pieter Abbeel}, {and} \bibinfo{person}{Sergey Levine}.} \bibinfo{year}{2019}\natexlab{}.
\newblock \showarticletitle{Mcp: Learning composable hierarchical control with multiplicative compositional policies}. In \bibinfo{booktitle}{\emph{Advances in Neural Information Processing Systems (NeurIPS)}}.
\newblock


\bibitem[Radford et~al\mbox{.}(2021)]%
        {radford2021learning}
\bibfield{author}{\bibinfo{person}{Alec Radford}, \bibinfo{person}{Jong~Wook Kim}, \bibinfo{person}{Chris Hallacy}, \bibinfo{person}{Aditya Ramesh}, \bibinfo{person}{Gabriel Goh}, \bibinfo{person}{Sandhini Agarwal}, \bibinfo{person}{Girish Sastry}, \bibinfo{person}{Amanda Askell}, \bibinfo{person}{Pamela Mishkin}, \bibinfo{person}{Jack Clark}, {et~al\mbox{.}}} \bibinfo{year}{2021}\natexlab{}.
\newblock \showarticletitle{Learning transferable visual models from natural language supervision}. In \bibinfo{booktitle}{\emph{International Conference on Machine Learning (ICML)}}.
\newblock


\bibitem[Savva et~al\mbox{.}(2016)]%
        {savva2016pigraphs}
\bibfield{author}{\bibinfo{person}{Manolis Savva}, \bibinfo{person}{Angel~X Chang}, \bibinfo{person}{Pat Hanrahan}, \bibinfo{person}{Matthew Fisher}, {and} \bibinfo{person}{Matthias Nie{\ss}ner}.} \bibinfo{year}{2016}\natexlab{}.
\newblock \showarticletitle{Pigraphs: learning interaction snapshots from observations}.
\newblock \bibinfo{journal}{\emph{ACM Transactions on Graphics (TOG)}} \bibinfo{volume}{35}, \bibinfo{number}{4} (\bibinfo{year}{2016}), \bibinfo{pages}{1--12}.
\newblock


\bibitem[Starke et~al\mbox{.}(2019)]%
        {starke2019neural}
\bibfield{author}{\bibinfo{person}{Sebastian Starke}, \bibinfo{person}{He Zhang}, \bibinfo{person}{Taku Komura}, {and} \bibinfo{person}{Jun Saito}.} \bibinfo{year}{2019}\natexlab{}.
\newblock \showarticletitle{Neural state machine for character-scene interactions}.
\newblock \bibinfo{journal}{\emph{ACM Transactions on Graphics (TOG)}} \bibinfo{volume}{38}, \bibinfo{number}{6} (\bibinfo{year}{2019}), \bibinfo{pages}{178}.
\newblock


\bibitem[Taheri et~al\mbox{.}(2022a)]%
        {taheri2022goal}
\bibfield{author}{\bibinfo{person}{Omid Taheri}, \bibinfo{person}{Vasileios Choutas}, \bibinfo{person}{Michael~J Black}, {and} \bibinfo{person}{Dimitrios Tzionas}.} \bibinfo{year}{2022}\natexlab{a}.
\newblock \showarticletitle{GOAL: Generating 4D Whole-Body Motion for Hand-Object Grasping}. In \bibinfo{booktitle}{\emph{Conference on Computer Vision and Pattern Recognition (CVPR)}}.
\newblock


\bibitem[Taheri et~al\mbox{.}(2022b)]%
        {taheri2021Goal}
\bibfield{author}{\bibinfo{person}{Omid Taheri}, \bibinfo{person}{Vasileios Choutas}, \bibinfo{person}{Michael~J. Black}, {and} \bibinfo{person}{Dimitrios Tzionas}.} \bibinfo{year}{2022}\natexlab{b}.
\newblock \showarticletitle{{GOAL}: {G}enerating {4D} Whole-Body Motion for Hand-Object Grasping}. In \bibinfo{booktitle}{\emph{Conference on Computer Vision and Pattern Recognition (CVPR)}}.
\newblock


\bibitem[Taheri et~al\mbox{.}(2020)]%
        {taheri2020grab}
\bibfield{author}{\bibinfo{person}{Omid Taheri}, \bibinfo{person}{Nima Ghorbani}, \bibinfo{person}{Michael~J Black}, {and} \bibinfo{person}{Dimitrios Tzionas}.} \bibinfo{year}{2020}\natexlab{}.
\newblock \showarticletitle{GRAB: A dataset of whole-body human grasping of objects}. In \bibinfo{booktitle}{\emph{European Conference on Computer Vision (ECCV)}}.
\newblock


\bibitem[Tevet et~al\mbox{.}(2022)]%
        {tevet2022human}
\bibfield{author}{\bibinfo{person}{Guy Tevet}, \bibinfo{person}{Sigal Raab}, \bibinfo{person}{Brian Gordon}, \bibinfo{person}{Yoni Shafir}, \bibinfo{person}{Daniel Cohen-or}, {and} \bibinfo{person}{Amit~Haim Bermano}.} \bibinfo{year}{2022}\natexlab{}.
\newblock \showarticletitle{Human Motion Diffusion Model}. In \bibinfo{booktitle}{\emph{International Conference on Learning Representations (ICLR)}}.
\newblock


\bibitem[Vaswani et~al\mbox{.}(2017)]%
        {vaswani2017attention}
\bibfield{author}{\bibinfo{person}{Ashish Vaswani}, \bibinfo{person}{Noam Shazeer}, \bibinfo{person}{Niki Parmar}, \bibinfo{person}{Jakob Uszkoreit}, \bibinfo{person}{Llion Jones}, \bibinfo{person}{Aidan~N Gomez}, \bibinfo{person}{{\L}ukasz Kaiser}, {and} \bibinfo{person}{Illia Polosukhin}.} \bibinfo{year}{2017}\natexlab{}.
\newblock \showarticletitle{Attention is all you need}. In \bibinfo{booktitle}{\emph{Advances in Neural Information Processing Systems (NeurIPS)}}.
\newblock


\bibitem[von Marcard et~al\mbox{.}(2018)]%
        {vonMarcard2018}
\bibfield{author}{\bibinfo{person}{Timo von Marcard}, \bibinfo{person}{Roberto Henschel}, \bibinfo{person}{Michael Black}, \bibinfo{person}{Bodo Rosenhahn}, {and} \bibinfo{person}{Gerard Pons-Moll}.} \bibinfo{year}{2018}\natexlab{}.
\newblock \showarticletitle{Recovering Accurate 3D Human Pose in The Wild Using IMUs and a Moving Camera}. In \bibinfo{booktitle}{\emph{European Conference on Computer Vision (ECCV)}}.
\newblock


\bibitem[Wang et~al\mbox{.}(2021a)]%
        {wang2021synthesizing}
\bibfield{author}{\bibinfo{person}{Jiashun Wang}, \bibinfo{person}{Huazhe Xu}, \bibinfo{person}{Jingwei Xu}, \bibinfo{person}{Sifei Liu}, {and} \bibinfo{person}{Xiaolong Wang}.} \bibinfo{year}{2021}\natexlab{a}.
\newblock \showarticletitle{Synthesizing Long-Term 3D Human Motion and Interaction in 3D Scenes}. In \bibinfo{booktitle}{\emph{Conference on Computer Vision and Pattern Recognition (CVPR)}}.
\newblock


\bibitem[Wang et~al\mbox{.}(2021b)]%
        {wang2021scene}
\bibfield{author}{\bibinfo{person}{Jingbo Wang}, \bibinfo{person}{Sijie Yan}, \bibinfo{person}{Bo Dai}, {and} \bibinfo{person}{Dahua Lin}.} \bibinfo{year}{2021}\natexlab{b}.
\newblock \showarticletitle{Scene-aware generative network for human motion synthesis}. In \bibinfo{booktitle}{\emph{Conference on Computer Vision and Pattern Recognition (CVPR)}}.
\newblock


\bibitem[Wang et~al\mbox{.}(2022)]%
        {wang2022humanise}
\bibfield{author}{\bibinfo{person}{Zan Wang}, \bibinfo{person}{Yixin Chen}, \bibinfo{person}{Tengyu Liu}, \bibinfo{person}{Yixin Zhu}, \bibinfo{person}{Wei Liang}, {and} \bibinfo{person}{Siyuan Huang}.} \bibinfo{year}{2022}\natexlab{}.
\newblock \showarticletitle{Humanise: Language-conditioned human motion generation in 3d scenes}. In \bibinfo{booktitle}{\emph{Advances in Neural Information Processing Systems (NeurIPS)}}.
\newblock


\bibitem[Wu et~al\mbox{.}(2022)]%
        {wu2022saga}
\bibfield{author}{\bibinfo{person}{Yan Wu}, \bibinfo{person}{Jiahao Wang}, \bibinfo{person}{Yan Zhang}, \bibinfo{person}{Siwei Zhang}, \bibinfo{person}{Otmar Hilliges}, \bibinfo{person}{Fisher Yu}, {and} \bibinfo{person}{Siyu Tang}.} \bibinfo{year}{2022}\natexlab{}.
\newblock \showarticletitle{SAGA: Stochastic Whole-Body Grasping with Contact}. In \bibinfo{booktitle}{\emph{European Conference on Computer Vision (ECCV)}}.
\newblock


\bibitem[Xiao et~al\mbox{.}(2024)]%
        {xiao2024unified}
\bibfield{author}{\bibinfo{person}{Zeqi Xiao}, \bibinfo{person}{Tai Wang}, \bibinfo{person}{Jingbo Wang}, \bibinfo{person}{Jinkun Cao}, \bibinfo{person}{Wenwei Zhang}, \bibinfo{person}{Bo Dai}, \bibinfo{person}{Dahua Lin}, {and} \bibinfo{person}{Jiangmiao Pang}.} \bibinfo{year}{2024}\natexlab{}.
\newblock \showarticletitle{Unified Human-Scene Interaction via Prompted Chain-of-Contacts}. In \bibinfo{booktitle}{\emph{International Conference on Learning Representations (ICLR)}}.
\newblock


\bibitem[Yi et~al\mbox{.}(2024)]%
        {yi2024generating}
\bibfield{author}{\bibinfo{person}{Hongwei Yi}, \bibinfo{person}{Justus Thies}, \bibinfo{person}{Michael~J Black}, \bibinfo{person}{Xue~Bin Peng}, {and} \bibinfo{person}{Davis Rempe}.} \bibinfo{year}{2024}\natexlab{}.
\newblock \showarticletitle{Generating Human Interaction Motions in Scenes with Text Control}. In \bibinfo{booktitle}{\emph{Conference on Computer Vision and Pattern Recognition (CVPR)}}.
\newblock


\bibitem[Zhang et~al\mbox{.}(2020b)]%
        {zhangsiwei2020generating}
\bibfield{author}{\bibinfo{person}{Siwei Zhang}, \bibinfo{person}{Yan Zhang}, \bibinfo{person}{Qianli Ma}, \bibinfo{person}{Michael~J Black}, {and} \bibinfo{person}{Siyu Tang}.} \bibinfo{year}{2020}\natexlab{b}.
\newblock \showarticletitle{Generating person-scene interactions in 3d scenes}. In \bibinfo{booktitle}{\emph{International Conference on 3D Vision (3DV)}}.
\newblock


\bibitem[Zhang et~al\mbox{.}(2022)]%
        {zhang2022couch}
\bibfield{author}{\bibinfo{person}{Xiaohan Zhang}, \bibinfo{person}{Bharat~Lal Bhatnagar}, \bibinfo{person}{Sebastian Starke}, \bibinfo{person}{Vladimir Guzov}, {and} \bibinfo{person}{Gerard Pons-Moll}.} \bibinfo{year}{2022}\natexlab{}.
\newblock \showarticletitle{Couch: Towards controllable human-chair interactions}. In \bibinfo{booktitle}{\emph{European Conference on Computer Vision (ECCV)}}.
\newblock


\bibitem[Zhang et~al\mbox{.}(2020a)]%
        {zhang2020generating}
\bibfield{author}{\bibinfo{person}{Yan Zhang}, \bibinfo{person}{Mohamed Hassan}, \bibinfo{person}{Heiko Neumann}, \bibinfo{person}{Michael~J Black}, {and} \bibinfo{person}{Siyu Tang}.} \bibinfo{year}{2020}\natexlab{a}.
\newblock \showarticletitle{Generating 3d people in scenes without people}. In \bibinfo{booktitle}{\emph{Conference on Computer Vision and Pattern Recognition (CVPR)}}.
\newblock


\bibitem[Zhang and Tang(2022)]%
        {WanderingsofOdysseusIn3DScenes}
\bibfield{author}{\bibinfo{person}{Yan Zhang} {and} \bibinfo{person}{Siyu Tang}.} \bibinfo{year}{2022}\natexlab{}.
\newblock \showarticletitle{The Wanderings of Odysseus in 3D Scenes}. In \bibinfo{booktitle}{\emph{Conference on Computer Vision and Pattern Recognition (CVPR)}}.
\newblock


\bibitem[Zhao et~al\mbox{.}(2022)]%
        {zhao2022compositional}
\bibfield{author}{\bibinfo{person}{Kaifeng Zhao}, \bibinfo{person}{Shaofei Wang}, \bibinfo{person}{Yan Zhang}, \bibinfo{person}{Thabo Beeler}, {and} \bibinfo{person}{Siyu Tang}.} \bibinfo{year}{2022}\natexlab{}.
\newblock \showarticletitle{Compositional human-scene interaction synthesis with semantic control}. In \bibinfo{booktitle}{\emph{European Conference on Computer Vision (ECCV)}}.
\newblock


\bibitem[Zhao et~al\mbox{.}(2023)]%
        {zhao2023synthesizing}
\bibfield{author}{\bibinfo{person}{Kaifeng Zhao}, \bibinfo{person}{Yan Zhang}, \bibinfo{person}{Shaofei Wang}, \bibinfo{person}{Thabo Beeler}, {and} \bibinfo{person}{Siyu Tang}.} \bibinfo{year}{2023}\natexlab{}.
\newblock \showarticletitle{Synthesizing Diverse Human Motions in 3D Indoor Scenes}. In \bibinfo{booktitle}{\emph{International Conference on Computer Vision (ICCV)}}.
\newblock


\end{thebibliography}

\begin{figure*}
    \includegraphics[width=\linewidth]{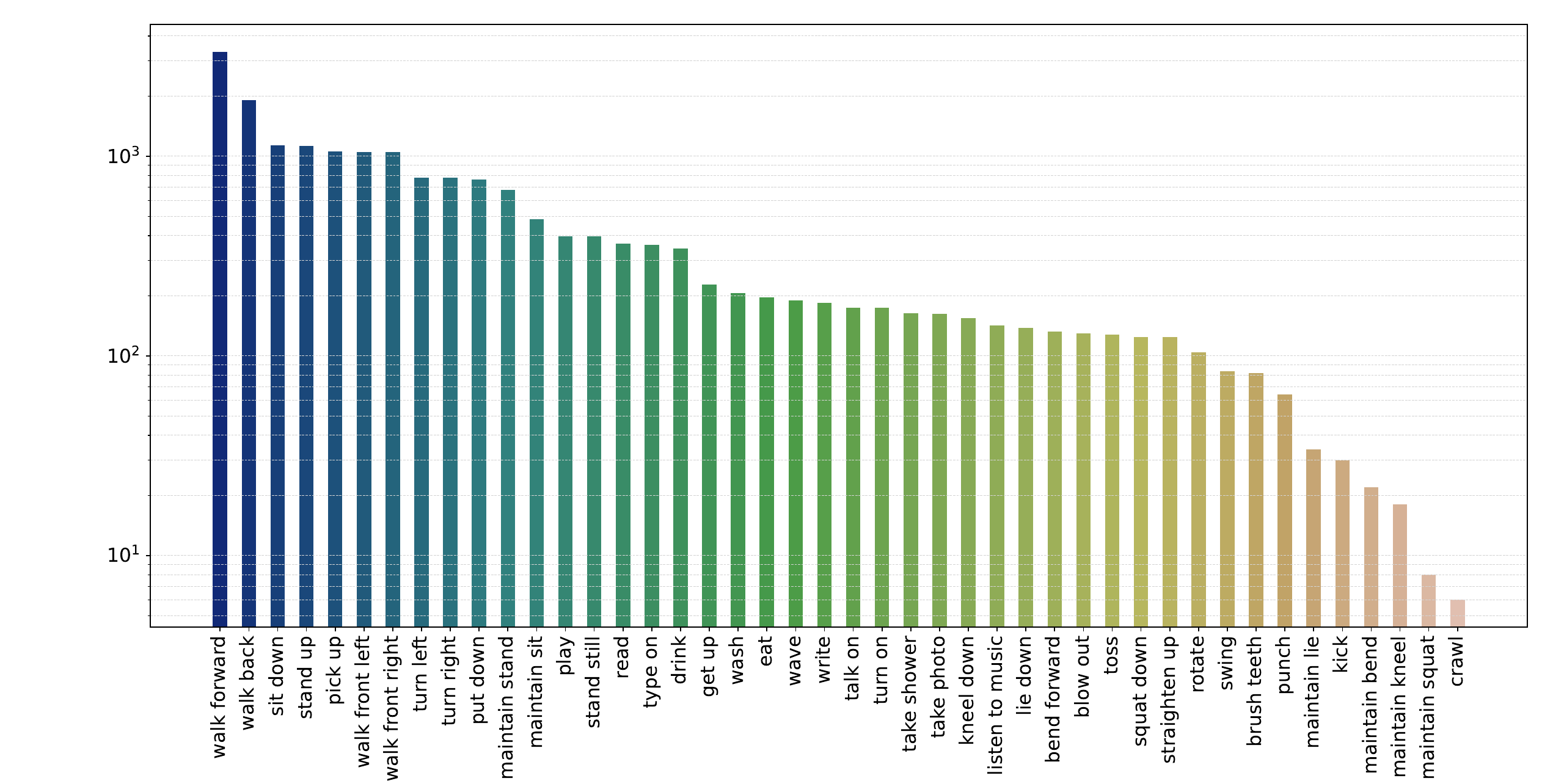}
    \caption{\textbf{Number of occurrences of each motion type in LINGO dataset.}\\ \hspace*{\fill} \\}
    \label{fig:motion_cnt}
\end{figure*}

\clearpage
\clearpage
\appendix
\renewcommand\thefigure{A\arabic{figure}}
\setcounter{figure}{0}
\renewcommand\thetable{A\arabic{table}}
\setcounter{table}{0}
\renewcommand\theequation{A\arabic{equation}}
\setcounter{equation}{0}
\pagenumbering{arabic}
\renewcommand*{\thepage}{A\arabic{page}}
\setcounter{footnote}{0}

\section{Implementation Details}

In this section, we describe the detailed architecture of each module in our framework, along with the training configurations.

\subsection{Motion Diffusion Module}

The motion diffusion module employs a Transformer encoder architecture \cite{vaswani2017attention} with 8 layers and 16 attention heads, which has proven highly effective in modeling sequential data. The input to the model consists of tokens representing the noised body joints and additional tokens for the condition information. To achieve auto-regressive generation, we fix the first two tokens of the current segment, copy the value of the last two frames from the last segment, and zero out the noise applied to them, during both training and sampling. In addition to the body joint tokens, four other tokens are introduced to incorporate the conditions, representing the scene, text, pelvis, and hand goal location, which provide crucial context for generating coherent and relevant motions. The details of these conditioning tokens will be discussed in the following three subsections. An embedding of diffusion timestep is added to the four conditioning tokens to incorporate temporal information into the model. All tokens undergo a positional encoding before being fed into the Transformer model.

\subsection{Scene Encoder}

The current scene voxel and its predictive counterpart are concatenated along the channel dimension, creating a unified 64-channel, 32x32 image. This image is segmented into 8x8 patches, which serve as input for a \ac{vit} \cite{dosovitskiy2020vit} consisting of 6 layers and 16 attention heads. The ViT processes these patches and produces a 512-dimensional feature vector. This vector is then used as the scene conditioning token in the motion diffusion module, ensuring context-aware motion synthesis.

When the target action involves interactions unrelated to the scene, such as drinking water from a bottle or talking on the phone, we mask the scene conditional token as all zeros. This approach is intended to prevent interference from scene information during the generation process of scene-independent interactions.

\subsection{Frame-embedded Text Encoder}

We employ the CLIP encoder \cite{radford2021learning} to convert raw text descriptions into 768-dimensional latent vectors. These vectors are then transformed into 512-dimensional vectors using an MLP model. Simultaneously, a sinusoidal positional encoder converts the frame number from an integer to a 512-dimensional vector, forming the frame embedding. We then add the text embedding to the frame embedding and pass the result through another MLP layer to obtain the final text conditional token, the input for our motion diffusion module.

During the sampling phase, at each step of autoregressive generation, the frame number input to the model increases as the number of generated frames increases. This aligns with the increased rate during training. In particular, by controlling the rate at which the frame number increases, we can adjust the total duration of the generated action. Due to the periodic nature of locomotion, such as walking, we set the frame number to zero during both the training and sampling processes for locomotion.

\subsection{Goal Encoder}

We train separate MLPs to embed locomotion and hand goals, resulting in embeddings for pelvic and hand goals, respectively.
For locomotion tasks, we remove the vertical component of the goal location and retain only the two-dimensional horizontal coordinates as input for the model. In object-reaching tasks, the target coordinates of the hand serve as input for the model. We mask the pelvis or hand goal tokens as all zeros for actions that do not involve locomotion or hand reaching.

\subsection{Autonomous Scheduler}

Our scheduler model generates a value ranging from 0 to 1, which indicates whether the previously generated motion clip has completed its entire semantic motion. This value subsequently determines whether the current motion clip should maintain the existing semantic or initiate the first motion clip of a new semantic. Our Scheduler model utilizes a Transformer encoder with 3 layers, 8 attention heads, and a hidden dimension of 512. We leverage a model identical to the structure described in \cref{sec:method_text_embed} to embed the current frame number and the given language instruction as a text conditioning token. This token, along with other motion frames, serves as the input to the Transformer encoder.

Due to the simplicity of this task, we train the scheduler model on the entire LINGO dataset for only 5 epochs. We use a batch size of 1024 and a learning rate of 0.0001, employing the Adam optimizer with its default parameter settings. The model converges effectively and demonstrates strong performance.

\subsection{Training Configuration}

The training of our motion diffusion module is conducted with a carefully selected set of hyperparameters to ensure optimal performance. We utilize a learning rate of 0.0001. The number of diffusion timesteps is fixed at 100, balancing computational feasibility and the quality of generated samples. In addition, we adopt a linear noise schedule, which gradually increases noise levels throughout the diffusion process. We use 4 NVIDIA A100 GPUs to train over 500 epochs with a batch size of 1024, ensuring sufficient exposure to the training data for convergence. These hyperparameter choices are informed by prior literature and empirical experimentation.

\begin{figure*}[t!]
    \includegraphics[width=\linewidth]{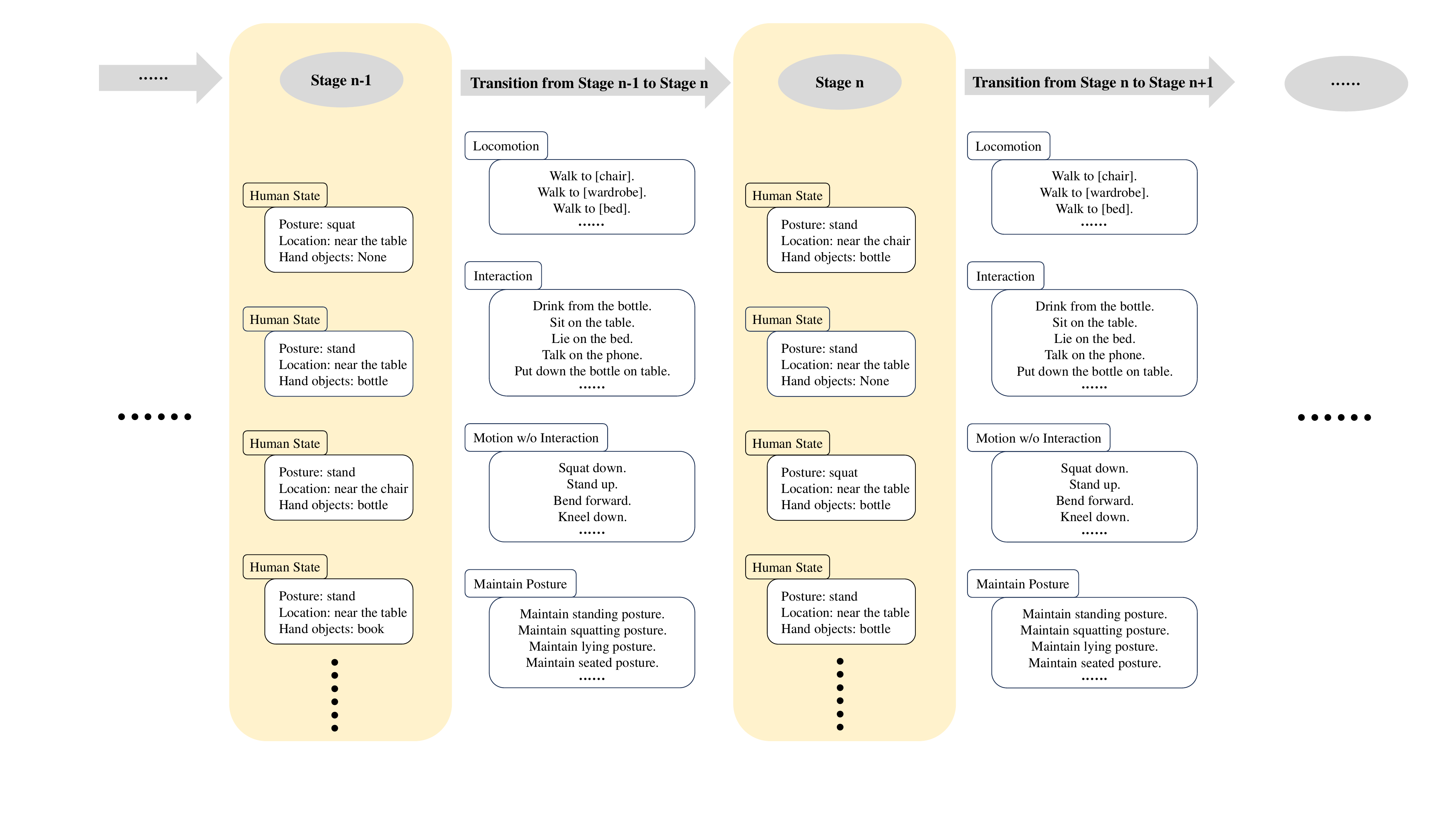}
    \caption{\textbf{Motion Planner.} A Markov Chain generates the next instruction to guarantee plausible interaction and maintain a balanced distribution of motion types. The Motion Planner provides language instructions to the Actor.}
    \label{fig:motion_planer}
\end{figure*}

\section{LINGO Dataset}

In this section, we elaborate on the recording process of the LINGO dataset and statistics of the LINGO dataset in detail.

\subsection{How LINGO Dataset Is Produced}

Producing a VR-assisted motion-captured dataset is a complex process that involves multiple people, specialized equipment, and custom software. In this section, we provide an overview of the key components and steps involved in this process.

\subsubsection{People Participants}

The MoCap process involves three main roles.

\paragraph{Actor} The Actor performs the motions while wearing a MoCap suit and a VR headset. 

\paragraph{Controller} The Controller provides the Actor with language descriptions of the motions to perform. 

\paragraph{Assistant} The Assistant marks the frames when the Actor picks up or puts down hand-held objects, ensuring accurate synchronization between the motion data and the object interactions.

\subsubsection{MoCap Add-on}

We designed a custom Blender add-on to facilitate the MoCap process. This addon has three main functions. First, it displays the live motion of the Actor in the physical environment, superimposed on the virtual scene. This allows the production team to identify and correct errors such as penetration between the Actor's virtual representation and the scene objects or erroneous motion capture data. Second, the addon shows the Actor's VR headset view, which helps the team adjust the motion capture setup to suit the current motion type and Actor best. Third, the add-on provides a third-person view that follows the Actor's movement, similar to a third-person video game camera. This helps the Controller give orientation-related locomotion instructions, such as ``walk to the right.'' The addon also links the VR and VICON systems, projecting rendered views and language instructions to the Actor and repeatedly aligning the real-world and virtual-world coordinates to maintain synchronization.

\subsubsection{Motion Planner Add-on}

Another custom addon, the Motion Planner (\cref{fig:motion_planer}), generates a sequence of instructions for the Actor to perform in the current scene as a Markov Chain. The input to the Motion Planner is a list of candidate interactions, their properties, and constraints. For example, some interactions may require the use of one or both hands or may have specific starting or ending positions. The Motion Planner considers these constraints and outputs a sequence of motion instructions that satisfy them. The Motion Planner also helps maintain a balanced distribution in the dataset. The Controller advances the instructions displayed in the Actor's VR headset by pressing a ``go to next instruction'' button.

\subsubsection{Preparation}

Before starting the MoCap process, the scene files are prepared in Blender. This involves selecting scene objects for the Actor to interact with and adding small interactable objects as needed. For sittable objects such as sofas and chairs, placeholders are placed in the physical environment to support the Actor during the capture session. The interaction types for each scene are specified and input to the Motion Planner. The VICON system is warmed up and calibrated to ensure accurate tracking. The VR headset is initialized by aligning the virtual world with the real world using a calibration procedure.

\subsubsection{Motion Capturing}

During the MoCap session, the Actor stands in a plausible location within the scene. The Controller checks that everything is ready and displays the first motion instruction in the Actor's headset. For interactive motions, such as picking up or manipulating objects, the Controller determines when the interaction is finished and advances to the next instruction. For grasping motions, the Assistant marks the time frames when the Actor grasps or releases the object, ensuring that these critical events are accurately recorded in the dataset and the object is attached to hands correctly. For locomotion, the Controller guides the Actor to the goal location using arrow keys, with specific direction-related information projected in the VR headset and recorded as annotations for the LINGO dataset.

During the MoCap process, the Actor performs the actions based on the language instructions projected on their VR headset. The VICON system tracks and records their body movements in real-time as the Actor moves and interacts within the physical space. Simultaneously, the captured motion data is instantly transmitted to the virtual scene, where the Actor's virtual body is rendered in real-time. This real-time reconstruction allows the Actor to see their virtual representation within the VR headset, creating a highly immersive and interactive experience. 

\subsubsection{Data Post-processing}

After the MoCap session, the raw motion capture data is split into segments according to motion types. The Motion Planner Add-on accompanies each segment with a raw text annotation describing the performed motion. To enhance the richness and variety of annotations, GPT-4 is used to augment the raw text into multiple versions, providing alternative descriptions and additional context. We also double the data size by mirroring both the motions and the annotations.

\begin{figure*}[t!]
    \includegraphics[width=\linewidth]{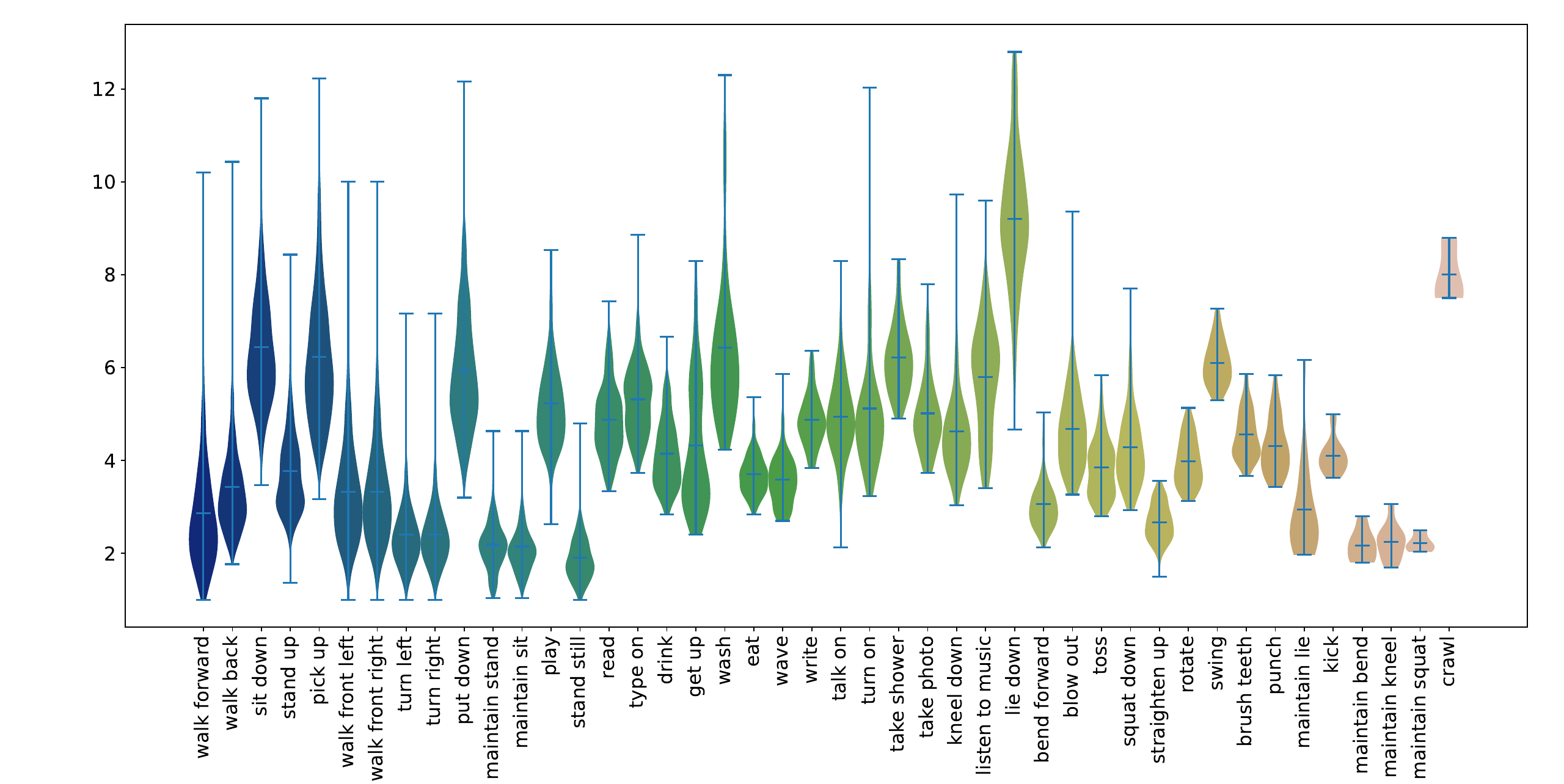}
    \caption{\textbf{Motion length distribution of each motion type in LINGO dataset.}}
    \label{fig:motion_len_violin}
\end{figure*}

\begin{figure}[t!]
    \includegraphics[width=\linewidth]{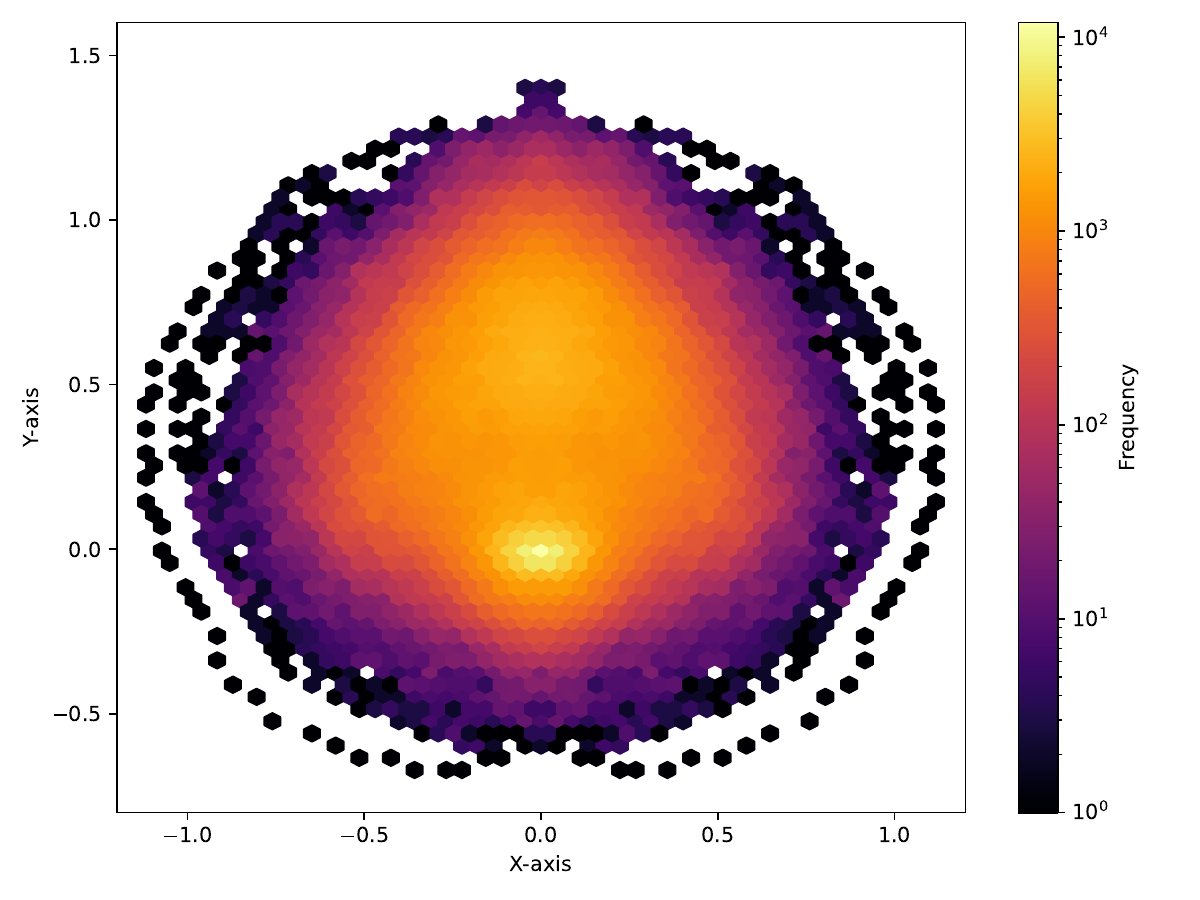}
    \caption{\textbf{Distribution of goal locations for all locomotion clips in the local coordinate system of the first frame.} The character is aligned to initially face the y-axis direction. Unit: meter.}
    \label{fig:walk_target}
\end{figure}

\subsection{Dataset Statistics}

\subsubsection{Interaction Types}

The LINGO dataset covers 40 types of motion listed in \cref{tab:motion_type}, including non-interactive motions such as locomotion and maintaining posture. For interactive motions, the LINGO dataset contains interaction with static scene objects (\eg, sit down, lie down) and small hand-held objects (\eg, cellphone, gamepad). \cref{fig:motion_cnt} counts the number of occurrences of each motion type, and \cref{fig:motion_len_violin} shows the distribution of the motion length. The detailed categorization is listed in \cref{tab:motion_type}.

\begin{table}[t!]
    \caption{\textbf{Motion types of LINGO.}}
    \label{tab:motion_type}
    \begin{tabularx}{\linewidth}{XX}
    \toprule
    \multicolumn{1}{l}{Motion Description}                           & Motion Name      \\
                                                                     \hline
    \multirow{4}{11em}{Move from one place to another by taking steps.} & walk forward  \\
                                                                     & walk back        \\
                                                                     & walk front left  \\
                                                                     & walk front right \\
                                                                     \hline
    \multirow{2}{11em}{Change the orientation of the body.}          & turn left        \\
                                                                     & turn right       \\
                                                                     \hline
    \multirow{2}{11em}{Change to a standing position.}                & stand up        \\
                                                                     & get up           \\
                                                                     \hline
    \multirow{16}{11em}{Interact with hand-held objects.}            & pick up          \\
                                                                     & put down         \\
                                                                     & take photo       \\
                                                                     & turn on          \\
                                                                     & write            \\
                                                                     & type on          \\
                                                                     & read             \\
                                                                     & play$^1$         \\
                                                                     & drink            \\
                                                                     & eat              \\
                                                                     & talk on          \\
                                                                     & listen to music  \\
                                                                     & brush teeth      \\
                                                                     & toss             \\
                                                                     & swing            \\
                                                                     & wave             \\
                                                                     \hline
    \multirow{2}{11em}{Pick up and put down hand-held objects.}      & pick up          \\
                                                                     & put down         \\
                                                                     \hline
    \multirow{6}{11em}{Stationary motions.}                          & stand still      \\
                                                                     & maintain lie     \\
                                                                     & maintain sit     \\
                                                                     & maintain bend    \\
                                                                     & maintain kneel   \\
                                                                     & maintain squat   \\
                                                                     \hline
    \multirow{5}{11em}{In-place motions.}                             & bend forward    \\
                                                                     & straighten up    \\
                                                                     & kneel down       \\
                                                                     & squat down       \\
                                                                     & crawl            \\
                                                                     \hline
    \multirow{9}{11em}{Interact with static scene objects.}           & sit down        \\
                                                                     & lie down         \\
                                                                     & punch            \\
                                                                     & kick             \\
                                                                     & wash             \\
                                                                     & take shower      \\
                                                                     & rotate           \\
                                                                     & play$^2$         \\
                                                                     & type             \\
                                                                     \bottomrule
    \end{tabularx}
    \footnotesize{$^1$ Play game and guitar. $^2$ Play drums and piano.}
\end{table}

\subsubsection{Locomotion Goal Distribution}

In the analysis of locomotion clips, we visualize the distribution of goal locations in \cref{fig:walk_target} represented in the canonical coordinate system of the first frame. This coordinate system is defined to align the character's initial orientation with the positive yaw direction. Using this consistent frame of reference, we can compare and study the relative positions of the goal locations across different clips. The plot represents the spatial distribution of the goal locations, with the origin (0, 0) corresponding to the character's starting position in the first frame. The plot's x-axis and y-axis represent the lateral and forward/backward directions, respectively, relative to the character's initial orientation. 

\subsubsection{Motion Length Distribution}

\cref{fig:motion_len_violin} presents the motion length distribution for various motion types in the LINGO dataset. Each violin represents a motion type, with the width of the violin indicating the density of data points at different motion lengths. The vertical axis measures the motion length, while the horizontal axis lists the motion types.

The motion lengths span from 1 to 12 seconds across all motion types. However, most of the data points lie between 2 and 6 seconds. The median motion length for most motion types is around 4-5 seconds. Some motions, like ``walk forward,'' are close to a normal distribution, while others, such as ``sit down'' or ``stand up,'' have a longer tail towards higher motion lengths due to the varied Actor preferences. Some motion types, such as ``lie,'' have significantly longer motions. Motions with respect to locomotion, such as ``walk forward'' and ``walk front left'' have similar distributions, while interactive motions have distinct distributions compared to the rest.

\subsubsection{Motion Occurance Count}

\cref{fig:motion_cnt} displays the number of occurrences for each motion type in the LINGO dataset. The vertical axis represents the count on a logarithmic scale, while the horizontal axis lists the various motion types.

The number of occurrences varies across motion types. Locomotion-related motions have the highest number of occurrences, exceeding 1000 instances in the dataset. This is because locomotion occurs between two interaction motions. Most interaction motion types, such as ``drink,'' ``eat,'' and ``wash,'' have around 200 occurrences or segments in the LINGO dataset.

\section{Task Planner}

We show a workable pipeline that leverages GPT4-o to break down complex instruction into sub-tasks.

\texttt{- prompt = ``I need you to help me complete a task now. I will give you a target action. The target action is:``+action+.'' You need to give in English a number of steps that I need to complete the target action. The steps should be as concise as possible without the need for irrelevant attributives. The details of the interaction action are not required. For example, there is no need to open the game console. The steps are divided into three categories: locomotion, grasp, and interaction, where locomotion only includes the movement of the person's position, Grasp only consists of the grabbing and touching of objects, and interaction includes people's operations on the appliance (such as listening to music with headphones, turning the door handle to open the door (excluding grabbing the door handle) )), indicate the hand when it comes to hand movements. Please complete this task according to some step examples I gave you. Example: ``+str(text\_list)+,'' output format: [\{"step" :,"step\_id":1,"category":\},...], only output the final format, no other nonsense''}

\texttt{- output = 
[
\{"step": "walk to the sofa", "step\_id": 1, "category": "locomotion"\},
\{"step": "sit down on the sofa", "step\_id": 2, "category": "locomotion"\},
\{"step": "pick up remote with left hand", "step\_id": 3, "category": "grasp"\},
\{"step": "turn on TV with left hand", "step\_id": 4, "category": "interaction"\},
\{"step": "watch TV", "step\_id": 5, "category": "interaction"\}
]}

\end{document}